%% file: main.tex
\documentclass[11pt]{article}

\usepackage[utf8]{inputenc} 
\usepackage[T1]{fontenc}    
\usepackage{geometry}       
\usepackage{setspace} 
\usepackage{lineno} 
\linenumbers

\usepackage{amsmath, amssymb, amsthm} 
\usepackage{algorithm}      
\usepackage{algorithmic}    

\usepackage{graphicx}       
\usepackage{booktabs}       
\usepackage{caption}        
\usepackage{subcaption}     
\usepackage{float}          
\usepackage[section]{placeins}

\usepackage[dvipsnames]{xcolor} 
\usepackage{hyperref}       
\hypersetup{
    colorlinks=true,
    linkcolor=NavyBlue,     
    filecolor=magenta,      
    urlcolor=Cyan,          
    citecolor=BrickRed,     
}

\usepackage{authblk}

\graphicspath{{figures/}}

\title{\textbf{Jacobian-Adaptive Weighting for Stability: Enhancing Long-term Rollout of Neural Partial Differential Equation Solvers via Spatially-Adaptive Regularization}}

\author[1]{Fengxiang Nie}
\author[1]{Yasuhiro Suzuki\thanks{Corresponding author. E-mail: suzukiy@hiroshima-u.ac.jp}}
\affil[1]{University of Hiroshima}

\begin{document}

\maketitle

\input{sections/00_abstract}
\input{sections/01_intro}
\input{sections/02_related}
\input{sections/03_method}
\input{sections/04_experiments}
\input{sections/05_discussion}
\input{sections/06_conclusion}

\section*{Declaration of Generative AI and AI-assisted technologies in the writing process}
During the preparation of this work, the authors used Google Gemini in order to improve the English language expression and to assist with the \LaTeX\ formatting. After using this tool/service, the authors reviewed and edited the content as needed and take full responsibility for the content of the publication.

\appendix
\input{sections/07_appendix}

\bibliographystyle{elsarticle-num} 
\bibliography{references} 

\end{document}

%% file: sections/00_abstract.tex
\begin{abstract}
Data-driven surrogate models can significantly accelerate the simulation of continuous dynamical systems, yet the step-wise accumulation of errors during autoregressive time-stepping often leads to spectral blow-up and unphysical divergence. Existing global regularization techniques can enforce contractive dynamics but uniformly damp high-frequency features, causing over-smoothing; meanwhile, long-horizon trajectory optimization methods are severely constrained by memory bottlenecks. This paper proposes Jacobian-Adaptive Weighting for Stability (JAWS), which reformulates operator learning as a Maximum A Posteriori (MAP) estimation problem with spatially heteroscedastic uncertainty, enabling the regularization strength to adapt automatically based on local physical complexity: enforcing contraction in smooth regions to suppress noise while relaxing constraints near singular features such as shocks to preserve gradient information. Experiments demonstrate that JAWS serves as an effective spectral pre-conditioner for trajectory optimization, allowing short-horizon, memory-efficient training to match the accuracy of long-horizon baselines. Validations on the 1D viscous Burgers' equation and 2D flow past a cylinder ($\text{Re}=400$ out-of-distribution generalization) confirm the method's advantages in long-term stability, preservation of physical conservation properties, and computational efficiency. This significant reduction in memory usage makes the method particularly well-suited for stable and efficient long-term simulation of large-scale flow fields in practical engineering applications.
\end{abstract}

\noindent\textbf{Keywords:} Neural operator, Autoregressive stability, Spatially-adaptive regularization, Heteroscedastic uncertainty, Spectral pre-conditioning, Computational fluid dynamics

%% file: sections/01_intro.tex
\section{Introduction}

Data-driven surrogate models have emerged as a new paradigm for traditional Computational Fluid Dynamics (CFD). By learning resolution-invariant mappings, methods such as the Fourier Neural Operator (FNO) \cite{li2020fourier} and DeepONet \cite{lu2021learning} significantly outperform conventional numerical solvers in computational efficiency. However, when these neural architectures are leveraged to construct autoregressive models for time-series predictions, their long-horizon rollouts are frequently hindered by instability. As noted by Brandstetter et al. \cite{brandstetter2022message}, the continuous accumulation of approximation errors during iterative rollouts leads to a severe distribution shift problem, which ultimately triggers unphysical divergent behaviors.

To mitigate this autoregressive instability, a straightforward approach is to enforce strict Lipschitz continuity across the entire network. Global regularization techniques based on this idea, such as Spectral Normalization \cite{miyato2018spectral}, mathematically enforce a global upper bound on the Lipschitz constant by constraining the spectral norm of the weight matrices, a property that can be utilized to ensure contractive dynamics in physical simulations. Yet, in the context of simulating physical systems, such global constraints uniformly damp high-frequency features, inevitably leading to severe over-smoothing that washes out critical physical details like sharp gradients and shocks. To guarantee asymptotic stability, the model must be globally contractive; however, capturing local high-frequency features requires the mapping to be locally expansive. This fundamental contradiction causes uniform constraints to often produce over-smoothed predictions lacking physical fidelity, resembling the effects of excessive artificial viscosity. 

To address the aforementioned limitations, this paper proposes a probabilistic regularization method named JAWS (Jacobian-Adaptive Weighting for Stability). JAWS formulates neural partial differential equation (PDE) solvers as a Maximum A Posteriori (MAP) estimation problem with spatially heteroscedastic uncertainty \cite{kendall2018multi}, the core idea being that the regularization strength is no longer fixed, but is automatically modulated by learnable spatial tolerance maps according to local physical complexity. In regions of shocks or steep gradients, the model automatically relaxes constraints to preserve key physical features; in smooth regions, it imposes strict constraints to suppress error accumulation. Furthermore, since JAWS effectively pre-conditions the operator's spectral properties, it allows extremely short-window trajectory optimization (Pushforward \cite{brandstetter2022message}) to match the long-term accuracy of long-window baselines, thereby breaking through the memory bottleneck of autoregressive training.

The main contributions of this paper can be summarized as follows:
\begin{itemize}
    \item \textbf{Theoretical Contribution:} We reveal the fundamental contradiction between traditional global Lipschitz constraints (e.g., spectral normalization) and the fidelity of local physical singularities such as shocks. Starting from the error propagation dynamics ($\boldsymbol{\epsilon}_{t+1} \approx \mathbf{J} \boldsymbol{\epsilon}_t$), we rigorously justify the necessity of spatially-adaptive regularization.
    \item \textbf{Methodological Contribution:} We propose JAWS (Jacobian-Adaptive Weighting for Stability), which, for the first time, reformulates operator learning as a MAP estimation problem with heteroscedastic uncertainty. By introducing learnable spatial tolerance maps $s_1(\mathbf{x})$ and $s_2(\mathbf{x})$, it achieves dynamic and intelligent modulation of regularization strength between smooth regions and singular features.
    \item \textbf{Engineering Contribution:} We design a hybrid training framework combining JAWS with extremely short-step trajectory optimization (Pushforward). Since JAWS serves as a spectral pre-conditioner that effectively suppresses high-frequency unstable modes, the model only needs to be trained within very short time windows (e.g., $k=5$ steps) with low memory consumption to match or even surpass the long-term rollout accuracy of traditional long-window ($k=10$ steps) baselines, completely breaking through the memory bottleneck of autoregressive training.
    \item \textbf{Experimental Contribution:} We comprehensively validate the superior performance of JAWS in long-term stability, high-frequency dynamics preservation, phase-space attractor retention, and computational efficiency on two classical nonlinear fluid dynamics benchmarks: the 1D viscous Burgers' equation (shock capturing) and 2D flow past a cylinder at Reynolds number $\text{Re}=400$ (out-of-distribution generalization).
\end{itemize}

In summary, this paper establishes a probabilistic regularization framework for data-driven surrogate models that transforms aleatoric uncertainty into a spatially-adaptive spectral modulation mechanism. Validations on the 1D viscous Burgers' equation and 2D flow past a cylinder demonstrate that the method not only accurately reproduces macroscopic physical statistics and stable attractor structures, but also exhibits reliable long-term tracking capabilities when confronted with out-of-distribution data. The low memory requirements and high computational efficiency of JAWS make it particularly suitable for practical engineering applications, such as real-time prediction and long-term stability analysis in large-scale flow field simulations.

%% file: sections/02_related.tex
\section{Related Work}
\label{sec:related}

\subsection{Data-Driven Surrogate Models and Long-Horizon Autoregressive Rollouts}
\label{sec:related_work_2_1}

In recent years, data-driven surrogate models have emerged as powerful tools to assist traditional CFD. Architectures such as the Fourier Neural Operator (FNO) \cite{li2020fourier} and DeepONet \cite{lu2021learning} offer resolution invariance and are significantly faster compared to conventional numerical solvers. By learning mappings between infinite-dimensional function spaces \cite{kovachki2023neural}, these models can accurately and efficiently capture complex physical phenomena. Despite their success in single-step or short-term predictions, these models face severe challenges when applied to long-horizon autoregressive rollouts. Autoregressive models frequently suffer from the problem of error accumulation. As pointed out by Brandstetter et al. \cite{brandstetter2022message}, the inputs during the iterative testing process gradually deviate from the single-step training distribution, which exacerbates the accumulation of approximation errors and inevitably leads to unphysical divergent behaviors in long-term trajectories.

\subsection{Stabilizing Dynamics: The Cost of Trajectory Optimization}
The mainstream strategy for mitigating the divergence problem is pushforward training (also known as time-bundling or trajectory optimization) \cite{brandstetter2022message}. By unrolling the solver for $k$ steps during training and minimizing the cumulative error $\sum_{i=1}^k \mathcal{L}(u_{t+i}, \hat{u}_{t+i})$, the model can explicitly learn to correct its own drift. Although this method is effective, its memory overhead is immense. Because Backpropagation Through Time (BPTT) requires storing the computational graphs for all $k$ steps, memory consumption grows linearly as $\mathcal{O}(k \cdot N)$, making long-horizon training unfeasible in high-resolution 3D simulations. This computational bottleneck motivates the search for alternative methods capable of \textit{implicitly} enforcing stability without the need for expensive unrolling.

\subsection{The Regularization Spectrum: Hard Constraints and Uniform Regularization}
Implicit stabilization methods typically constrain the model's Jacobian matrix $\mathbf{J} = \partial u_{t+1}/\partial u_t$ to ensure contractive dynamics ($\rho(\mathbf{J}) \le 1$). Existing approaches fall into two categories, each suffering from distinct issues:
\begin{itemize}
    \item \textbf{Hard Constraints (Spectral Normalization):} Methods like Spectral Normalization \cite{miyato2018spectral} force a global upper bound on the Lipschitz constant. While this guarantees stability, it often leads to severe \textbf{over-smoothing} of sharp discontinuities \cite{finlay2020train}. Functionally, its effect is akin to imposing excessive \textit{artificial viscosity}.
    \item \textbf{Uniform Jacobian Regularization:} In the broader field of deep learning, penalizing the Frobenius norm of the Jacobian matrix has been proven to improve generalization and robustness against adversarial attacks \cite{sokolic2017robust}. However, these methods typically apply a uniform penalty across the entire computational domain, failing to account for the spatially heterogeneous stability requirements in physical systems (e.g., the difference between a smooth region and a shock front).
\end{itemize}
As noted above, current regularization methods tend to swing to two extremes: hard constraints result in physical features being severely smoothed out, while uniform regularization fails to account for spatial heterogeneity. This work aims precisely to bridge the gap between these two extremes by proposing a \textit{spatially-adaptive} constraint mechanism---one that intelligently modulates the regularization strength according to the complexity of local flow features.

\subsection{Uncertainty Quantification and Spatially-Adaptive Regularization}
Uncertainty Quantification (UQ) in scientific machine learning has primarily focused on error estimation and active learning \cite{psaros2023uncertainty}. The methodological foundation of our approach is derived from Bayesian deep learning, specifically the use of homoscedastic uncertainty for multi-task loss weighting \cite{kendall2018multi}. We generalize this intuition to scientific computing by interpreting the learned variance parameters not merely as error estimates, but as a \textbf{spatially-adaptive spectral attention mechanism}.

This connects our method to the concepts of \textbf{shock-capturing schemes} or \textbf{adaptive artificial viscosity} in classical numerical analysis. In traditional high-resolution schemes (such as Weighted Essentially Non-Oscillatory (WENO)), numerical strategies are dynamically adjusted based on local flow features: numerical dissipation is increased near discontinuities to suppress unphysical Gibbs oscillations, while high-order schemes are employed in smooth regions to maintain computational accuracy. JAWS inherits this core philosophy of \textit{spatial adaptivity}, but realizes a complementary modulation direction in the regularization domain: it enforces strict Jacobian contraction constraints in smooth regions to suppress error accumulation, while relaxing constraints near shock fronts to prevent over-regularization from destroying sharp physical features. This uncertainty-based spatially-adaptive regularization approach remains largely unexplored in scientific machine learning. Our method not only provides a new perspective for understanding and designing stable autoregressive models, but also opens new directions for future research combining uncertainty quantification with physical constraints.

%% file: sections/03_method.tex
\section{Methodology}
\label{sec:method}

In this section, we derive the mathematical formulation of JAWS. We first establish the mathematical description of the sequence learning problem (\S\ref{subsec:problem}), and then theoretically justify the necessity of spatially-adaptive regularization by analyzing the error propagation dynamics (\S\ref{subsec:theory}). Based on this motivation, we present the Bayesian derivation of the JAWS objective function (\S\ref{subsec:jaws_derivation}) and its scalable implementation via stochastic trace estimation (\S\ref{subsec:hutchinson}). Finally, we describe how JAWS serves as an effective spectral pre-conditioner for long-horizon trajectory optimization (\S\ref{subsec:method_synergy}).

\subsection{Problem Formulation}
\label{subsec:problem}

Consider a continuous dynamical system governed by a time-dependent partial differential equation (PDE) defined on a spatial domain $\Omega \subset \mathbb{R}^d$ and a time interval $[0, T]$. Let $\mathcal{U}$ be a suitable function space taking values in $\mathbb{R}^{d_u}$. The system's temporal evolution is described by a non-linear operator $\mathcal{G}$:
\begin{equation}
    \frac{\partial \mathbf{u}}{\partial t} = \mathcal{G}(\mathbf{u}, \nabla \mathbf{u}, \dots), \quad \mathbf{u}(t) \in \mathcal{U}
\end{equation}
In the discrete setting, we aim to learn a data-driven surrogate model $\mathcal{M}_\theta$ that approximates the underlying operator between function spaces \cite{kovachki2023neural}. The goal is to approximate the transition dynamics (i.e., the flow map) such that the autoregressive rollout:
\begin{equation}
    \hat{\mathbf{u}}_{t+1} = \mathcal{M}_\theta(\hat{\mathbf{u}}_t), \quad \hat{\mathbf{u}}_0 = \mathbf{u}_0
\end{equation}
remains stable over long integration horizons and stays faithful to the ground truth trajectory $\{\mathbf{u}_t\}_{t=0}^T$.

\subsection{Theoretical Motivation}
\label{subsec:theory}

To understand the mechanism of long-term divergence, we examine the error propagation dynamics. Let $\boldsymbol{\epsilon}_t = \hat{\mathbf{u}}_t - \mathbf{u}_t$ denote a small perturbation at time $t$. Linearizing the surrogate model's dynamics around the true state $\mathbf{u}_t$ yields:
\begin{equation}
    \label{eq:error_dynamics}
    \boldsymbol{\epsilon}_{t+1} \approx \mathbf{J}(\mathbf{u}_t) \cdot \boldsymbol{\epsilon}_t, \quad \text{where } \mathbf{J}(\mathbf{u}_t) = \frac{\partial \mathcal{M}_\theta}{\partial \mathbf{u}}\bigg|_{\mathbf{u}_t}
\end{equation}
Equation \ref{eq:error_dynamics} reveals that strict numerical stability requires the local Jacobian $\mathbf{J}$ to be contractive, meaning its spectral radius must be bounded, $\rho(\mathbf{J}) \le 1$. Global regularization methods, such as Spectral Normalization, typically enforce a uniform constraint on the matrix norm (e.g., $\|\mathbf{J}\|_2 < 1$) across the entire spatial domain. However, this uniform constraint introduces a fundamental contradiction:
\begin{itemize}
    \item \textbf{Global Stability Requirement:} Suppressing the exponential amplification of numerical noise ($\boldsymbol{\epsilon}_t \to 0$) necessitates a strictly contractive Jacobian ($\|\mathbf{J}\| < 1$).
    \item \textbf{Local Fidelity Requirement:} Physical phenomena such as shock formation and turbulence cascades are characterized by extreme local gradients ($\|\nabla \mathbf{u}\| \gg 1$) and structural high-frequency energy. Enforcing a uniform upper bound on the operator's Lipschitz constant indiscriminately damps these high-frequency modes.
\end{itemize}
Consequently, a globally uniform constraint forces the model to violate the local fidelity requirement, functionally acting as excessive \textit{artificial viscosity} that leads to over-smoothed predictions. Conversely, completely unconstrained training violates the stability requirement, causing high-frequency aliasing errors to accumulate and eventually trigger spectral blow-up. This dilemma necessitates a \textit{spatially-adaptive} regularization constraint: one that strictly enforces contraction in smooth regions to suppress numerical noise, while selectively relaxing the constraint in structurally complex regions to preserve physical discontinuities.

\subsection{JAWS: Spatially-Adaptive Regularization via MAP Estimation}
\label{subsec:jaws_derivation}

Traditional global Jacobian penalties (or spectral normalization) are mathematically equivalent to imposing a uniform Gaussian prior on the operator's smoothness, which inherently violates the spatial heterogeneity of physical fields. Therefore, we reformulate the operator learning process as a Maximum A Posteriori (MAP) estimation problem with heteroscedastic uncertainty \cite{kendall2018multi}. We introduce a lightweight auxiliary network $\mathcal{H}_\phi(\mathbf{u}_t)$ (implemented as a lightweight convolutional prediction head on the backbone network's deep features, with negligible parameter count and inference latency) to output two spatially-varying tolerance fields (log-variance maps): $s_1(\mathbf{x})$ and $s_2(\mathbf{x})$.

\textbf{1. The Data Likelihood (Reconstruction).} 
We assume that the local model prediction error follows an independent Gaussian distribution with variance $\sigma_1^2(\mathbf{x}) = e^{s_1(\mathbf{x})}$ at each spatial location $\mathbf{x}$. Given the predicted state $\hat{\mathbf{u}}$ and the target state $\mathbf{y}$, the data likelihood is:
\begin{equation}
    p(\mathbf{y} | \hat{\mathbf{u}}, s_1) \propto \exp \left( - \frac{\|\mathbf{y} - \hat{\mathbf{u}}\|^2}{2 e^{s_1}} - \frac{1}{2}s_1 \right)
\end{equation}
Here, $s_1(\mathbf{x})$ captures the \textit{aleatoric uncertainty} of the prediction. In highly convective or hard-to-fit regions, the model can autonomously increase $s_1$ to attenuate the local loss weight, thereby avoiding overfitting to numerical noise.

\textbf{2. The Spatially-Adaptive Stability Prior.} 
Next, instead of employing hard constraints, we define the dynamic stability requirement as a spatially-heteroscedastic Gaussian prior on the Frobenius norm of the local Jacobian $\mathbf{J}(\mathbf{x})$, with variance $\sigma_2^2(\mathbf{x}) = e^{s_2(\mathbf{x})}$:
\begin{equation}
    p(\mathbf{J} | s_2) \propto \exp \left( - \frac{\|\mathbf{J}(\mathbf{x})\|_F^2}{2 e^{s_2}} - \frac{1}{2}s_2 \right)
\end{equation}
The term $e^{s_2(\mathbf{x})}$ acts as a Learnable Tolerance, granting the model the flexibility to locally violate the strict contraction rule:
\begin{itemize}
    \item \textbf{Near Shocks or Discontinuities:} Enforcing strict contraction would smooth out gradients and incur massive reconstruction errors. To minimize the total loss, the model is forced to increase $s_2(\mathbf{x})$, thereby relaxing the penalty on the Jacobian norm and allowing the local operator to ``expand'' to preserve high-frequency features.
    \item \textbf{In Smooth Regions:} When the data is well-fitted, the model decreases $s_2(\mathbf{x})$ to minimize the uncertainty penalty. This imposes an extremely strict penalty ($\|\mathbf{J}\| \to 0$), guaranteeing absolute numerical contraction and stability in these areas.
\end{itemize}

\textbf{3. The Joint MAP Objective.} 
According to Bayes' theorem, maximizing the posterior probability (MAP) is equivalent to minimizing the sum of the negative log-likelihood (NLL) and the negative log-prior. Using the log-variance parameterization to ensure numerical stability during optimization, we arrive at the final JAWS objective function:
\begin{equation}
    \label{eq:jaws_final}
    \mathcal{L}_{\text{JAWS}} = \sum_{\mathbf{x} \in \Omega} \bigg(
    \underbrace{\frac{1}{2}e^{-s_1} \|\mathbf{u}_{t+1} - \hat{\mathbf{u}}_{t+1}\|^2}_{\text{Adaptive Reconstruction}} 
    + \underbrace{\frac{1}{2}e^{-s_2} \|\mathbf{J}(\mathbf{x})\|_F^2}_{\text{Adaptive Regularization}}
    + \underbrace{\frac{1}{2}(s_1 + s_2)}_{\text{Complexity Penalty}} \bigg)
\end{equation}
This formulation elegantly achieves an adaptive balance. The final term, $\frac{1}{2}(s_1 + s_2)$, acts as a natural complexity regularizer that prevents the uncertainty variances from growing to infinity (i.e., avoiding the trivial solution where $s_1, s_2 \to \infty$). The model is compelled to find an optimal equilibrium between the cost of relaxing the tolerance and the benefit of fitting complex flow fields.

\subsection{Efficient Stochastic Estimation via Hutchinson's Trick}
\label{subsec:hutchinson}

For data-driven models operating on high-resolution spatial grids, the exact computation of the Jacobian Frobenius norm $\|\mathbf{J}\|_F^2$ is computationally intractable. It scales as $\mathcal{O}(N^2)$ or necessitates $N$ independent backpropagation passes, which is unacceptable for practical fluid dynamics surrogates. To make the JAWS objective scalable, we employ the Hutchinson trace estimator \cite{hutchinson1989stochastic}.

Using the algebraic identity $\|\mathbf{J}\|_F^2 = \text{Tr}(\mathbf{J}^T \mathbf{J})$, we introduce a random probe vector $\mathbf{v} \in \mathbb{R}^N$ sampled from a Rademacher distribution (entries $\pm 1$ with equal probability). The estimator is derived as:
\begin{equation}
    \begin{aligned}
    \mathbb{E}_{\mathbf{v}} [ \|\mathbf{J}\mathbf{v}\|_2^2 ] 
    &= \mathbb{E}_{\mathbf{v}} [ \mathbf{v}^T \mathbf{J}^T \mathbf{J} \mathbf{v} ] \\
    &= \text{Tr}(\mathbf{J}^T \mathbf{J} \mathbb{E}_{\mathbf{v}}[\mathbf{v}\mathbf{v}^T]) = \|\mathbf{J}\|_F^2
    \end{aligned}
\end{equation}
In practice, we approximate this mathematical expectation using a single random sample per training iteration. Crucially, the term $\mathbf{J}\mathbf{v}$ can be computed extraordinarily efficiently using Vector-Jacobian Products (VJP) via modern automatic differentiation frameworks:
\begin{equation}
    \mathbf{J}\mathbf{v} = \nabla_{\mathbf{u}} (\mathcal{M}_\theta(\mathbf{u}) \cdot \mathbf{v})
\end{equation}
This stochastic approach reduces the computational complexity to an $\mathcal{O}(1)$ backpropagation pass, adding virtually negligible memory and temporal overhead compared to standard unregularized training.

\textbf{Estimator Variance Note:} Although using a single random Rademacher vector $v \in \{-1, 1\}^N$ introduces considerable variance into the trace estimate for any individual forward pass, it is extremely efficient. In the deep learning context, this single-sample stochasticity is implicitly smoothed over thousands of training iterations by momentum-based optimizers (e.g., Adam). Furthermore, it has been theoretically proven that for a fixed matrix, the Rademacher distribution achieves the minimum variance among all unbiased stochastic trace estimators. Consequently, the model can stably converge to the optimal uncertainty parameters $s_1$ and $s_2$ without the need for computationally expensive multi-sample averaging.

\subsection{Synergy: Spectral Pre-conditioning for Trajectory Optimization}
\label{subsec:method_synergy}

While the single-step JAWS objective guarantees local Lipschitz boundedness, mitigating the accumulation of low-frequency phase drift over extended horizons still benefits from trajectory optimization (also known as Pushforward training \cite{stachenfeld2022learned}). However, pure Pushforward training over long sequences inevitably encounters optimization pathologies: the exponential divergence of gradients in chaotic systems leads to ill-conditioned Hessians, while Backpropagation Through Time (BPTT) consumes an excessive amount of memory.

To overcome these computational bottlenecks, we propose a hybrid optimization strategy where JAWS functions as a Spectral Pre-conditioner. We construct a composite loss over a short rollout window $K$:
\begin{equation}
    \mathcal{L}_{\text{Total}} =
    \mathcal{L}_{\text{JAWS}}(\mathbf{u}_t, \hat{\mathbf{u}}_{t+1}) 
    + \lambda \sum_{k=2}^K \|\mathbf{u}_{t+k} - \hat{\mathbf{u}}_{t+k}\|^2
\end{equation}

\textbf{Asymmetry of the Composite Loss Function:} The JAWS regularization is applied only to the base transition step ($k=1$) in Equation~(9). This asymmetry is a carefully designed topological structure intended to prevent uncertainty contamination. Applying the Jacobian penalty to subsequent unrolled steps ($k \ge 2$) would entangle the local spatial uncertainty estimates ($s_1, s_2$) with the chaotic temporal noise from backpropagated cumulative phase errors. By strictly bounding the local Lipschitz constant of the base step, JAWS effectively pre-conditions the operator's spectrum. This allows the subsequent Pushforward unrolling to focus on correcting macroscopic-scale drift without being disturbed by multi-scale gradient pathologies. For the specific implementation of this feature-level gradient topology isolation (including the structured information bottleneck design for uncertainty parameterization), see Appendix~\ref{sec:appendix_engineering}.

Gradient Detachment Strategy. 
A critical architectural detail for this synergy is the management of gradient flows. We apply a gradient detachment operation (\texttt{stop\_gradient}) to the state tensor before it enters the subsequent Pushforward rollouts:
\begin{equation}
    \hat{\mathbf{u}}_{t+1}^{\text{detach}} = \text{StopGrad}(\hat{\mathbf{u}}_{t+1})
\end{equation}
The multi-step Pushforward loss is then computed on the trajectory branching from $\hat{\mathbf{u}}_{t+1}^{\text{detach}}$. This decoupling mechanism ensures two vital properties:
\begin{enumerate}
    \item \textbf{Isolation of Uncertainty Estimation:} The spatially-adaptive tolerance maps $s_1$ and $s_2$ are optimized \textit{exclusively} against the high-fidelity, single-step physical transition. They are protected from being polluted by ill-conditioned gradient noise propagated backwards from long-term accumulated errors.
    \item \textbf{Well-Conditioned Base Operator:} The JAWS objective strictly conditions the Jacobian of the foundational step ($\rho(\mathbf{J}) \le 1$ in smooth regions). By relieving the Pushforward module of the burden of suppressing high-frequency instabilities, the trajectory optimization can exclusively focus on correcting low-frequency drift.
\end{enumerate}
By pre-conditioning the operator's spectral properties in advance, this synergistic mechanism dramatically enhances the model's training efficiency, enabling short-window, memory-efficient training to achieve long-term prediction accuracy comparable to or even better than long-window baseline methods.

%% file: sections/04_experiments.tex
﻿\section{Experiments}
\label{sec:experiments}

We empirically evaluate JAWS in two physical scenarios with distinct dimensions and dynamical characteristics. First, we test it on the 1D viscous Burgers' equation (\S\ref{subsec:1d_burgers}), a classic non-linear benchmark that perfectly illustrates the multi-scale interactions between smooth transport and shock formation. Second, we assess the model on strongly nonlinear 2D flow past a cylinder (\S\ref{subsec:2d_ns}), focusing on its out-of-distribution (OOD) extrapolation capabilities at a high Reynolds number ($\text{Re}=400$). Overall, these experiments fully demonstrate the core advantages of spatially-adaptive Jacobian regularization in guaranteeing long-term stability, maintaining physical conservation properties, and accurately capturing underlying dynamical invariants.

\subsection{1D Viscous Burgers' Equation: Shock Capturing and Stability}
\label{subsec:1d_burgers}

\subsubsection{Experimental Setup and Baselines}
\label{subsubsec:1d_setup}
We simulate the 1D viscous Burgers' equation, given by $u_t + u \cdot u_x = \nu \, u_{xx}$, with the kinematic viscosity $\nu \in [0.005, 0.02]$ and a time step of $\Delta t = 0.01$. The dataset is generated using a pseudo-spectral solver and comprises 2000 training trajectories and 500 testing trajectories (up to 400 steps per trajectory) under periodic boundary conditions. To ensure a fair evaluation of the regularization strategies, all compared methods share the exact same 1D periodic convolutional backbone.

We benchmark against the following constraints: (1) Baseline, trained with unconstrained Mean Squared Error (MSE); (2) Spectral Normalization~\cite{miyato2018spectral}, strictly enforcing a hard global Lipschitz constraint layer-by-layer; (3) JAWS-G, our global ablation variant that applies uniform Jacobian constraints; and (4) JAWS-S, our proposed spatially-adaptive Jacobian regularization method.

\subsubsection{Solution Accuracy Verification}
\label{subsubsec:1d_solution}

Before delving into the stability metrics, the most important step is to first verify whether the models can correctly solve the Burgers' equation. This is precisely the fundamental significance of choosing the Burgers' equation as a benchmark: it possesses analytically tractable nonlinear behavior (the competition between shock formation and viscous dissipation), allowing the model's basic solving capability to be assessed through direct comparison with high-accuracy pseudo-spectral numerical solutions.

\begin{figure}[htbp]
    \centering
    \includegraphics[width=\linewidth]{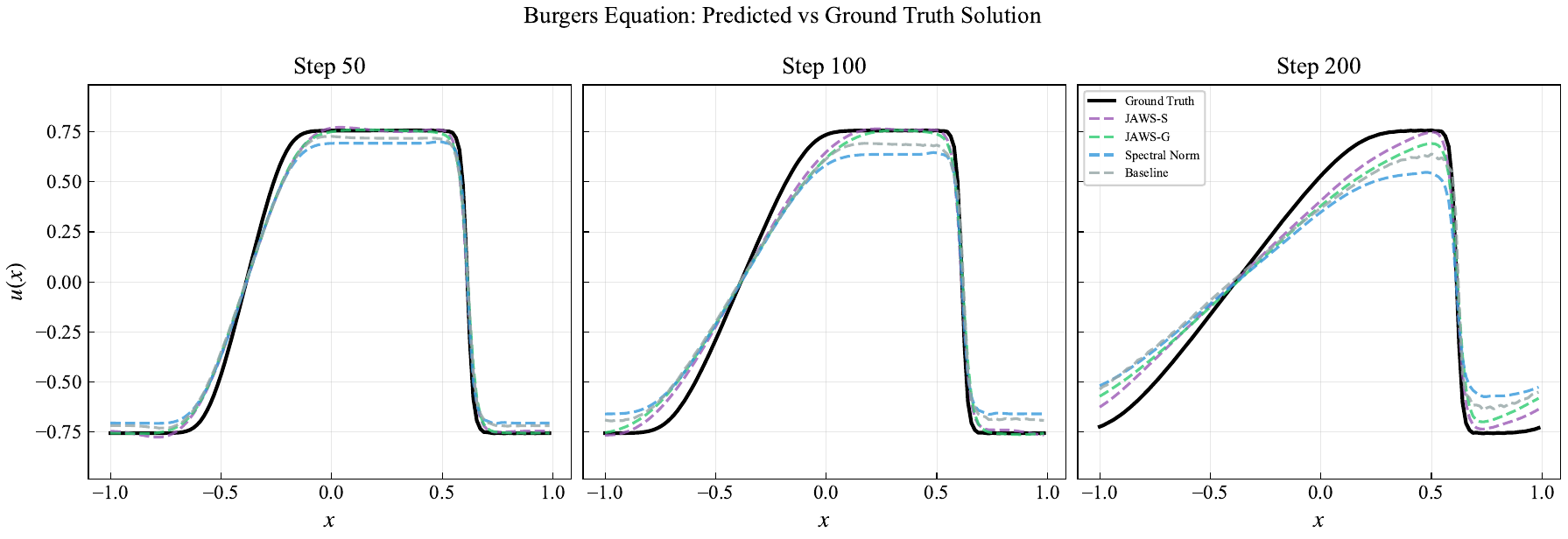}
    \caption{Burgers' equation solution verification: comparison of each model's predicted solution field $u(x)$ against pseudo-spectral ground truth at autoregressive rollout steps 50, 100, and 200 on the shock trajectory (quantitative errors in Table~\ref{tab:solution_accuracy}).}
    \label{fig:1d_solution_snapshots}
\end{figure}

Figure~\ref{fig:1d_solution_snapshots} shows the autoregressive predictions of each model on a shock initial condition trajectory. We select three representative time steps (step 50, step 100, step 200) and directly compare each model's predicted solution field $u(x)$ against the ground truth computed by the pseudo-spectral solver.

Table~\ref{tab:solution_accuracy} summarizes the relative $L^2$ errors of each model at three time steps. At step 50, all models achieve errors within the $7\%$--$10\%$ range, indicating that each method's single-step prediction accuracy is sufficient to correctly reproduce the spatial distribution of the solution. As the autoregressive rollout progresses to step 100, differences begin to emerge: JAWS-S's error only increases to $9.8\%$, while Spectral Norm and Baseline deteriorate to $17.4\%$ and $14.8\%$, respectively. By step 200, the divergence becomes even more dramatic---Spectral Norm and Baseline reach $30.3\%$ and $25.0\%$ respectively, while JAWS-S remains at only $14.9\%$, faithfully preserving the overall structure and shock features of the solution.

\begin{table}[htbp]
    \centering
    \caption{Relative $L^2$ error (\%) of each model at different autoregressive steps on the shock trajectory.}
    \label{tab:solution_accuracy}
    \begin{tabular}{lcccc}
        \toprule
        Step & JAWS-S & JAWS-G & Spectral Norm & Baseline \\
        \midrule
        50  & \textbf{7.3}  & 7.6  & 10.0 & 9.1 \\
        100 & \textbf{9.8}  & 12.3 & 17.4 & 14.8 \\
        200 & \textbf{14.9} & 20.3 & 30.3 & 25.0 \\
        \bottomrule
    \end{tabular}
\end{table}

These results directly verify two key facts: (1)~all models can correctly solve the Burgers' equation during short-term rollouts, with good consistency between predicted solution fields and ground truth; (2)~as the number of autoregressive steps increases, the stability differences among methods rapidly manifest---JAWS-S's spatially-adaptive Jacobian regularization effectively suppresses super-linear error growth, laying a reliable foundation for the subsequent quantitative stability analysis.

\subsubsection{Long-Term Stability and Spectral Maintenance}
\label{subsubsec:1d_stability}

\begin{figure}[htbp]
    \centering
    \begin{subfigure}[c]{0.48\linewidth}
        \centering
        \includegraphics[width=\linewidth]{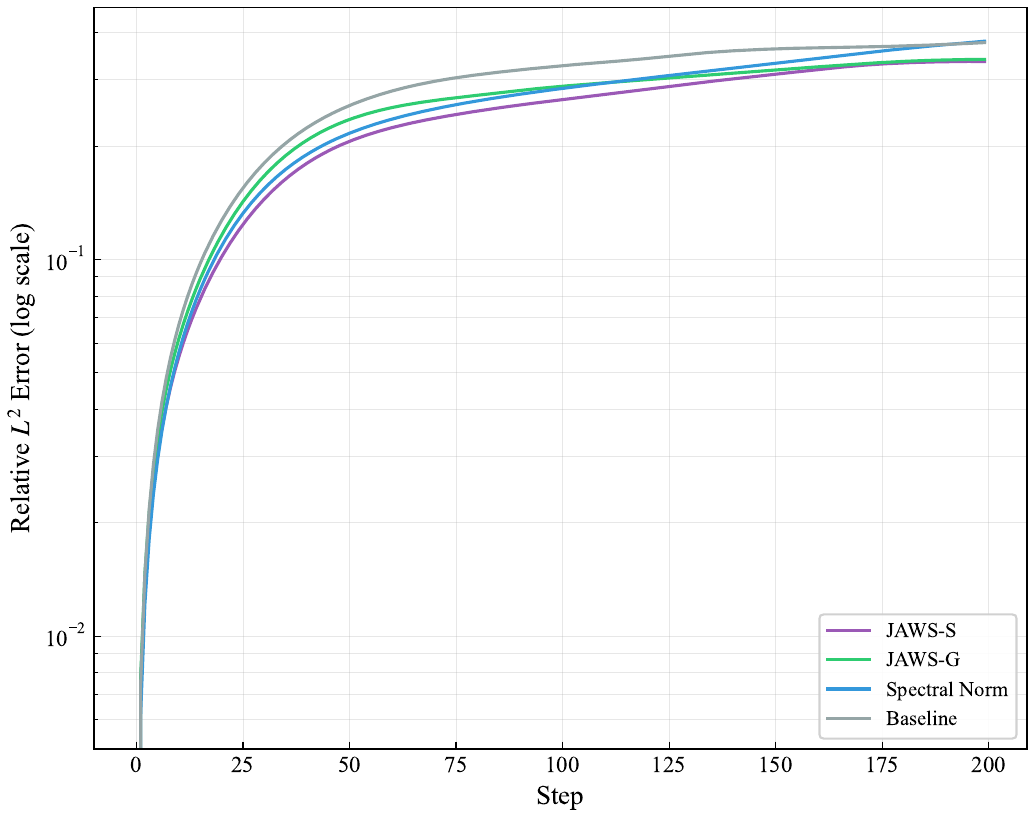}
        \caption{Relative $L^2$ Error Curve}
        \label{fig:1d_longterm_l2}
    \end{subfigure}
    \hfill
    \begin{subfigure}[c]{0.48\linewidth}
        \centering
        \includegraphics[width=\linewidth]{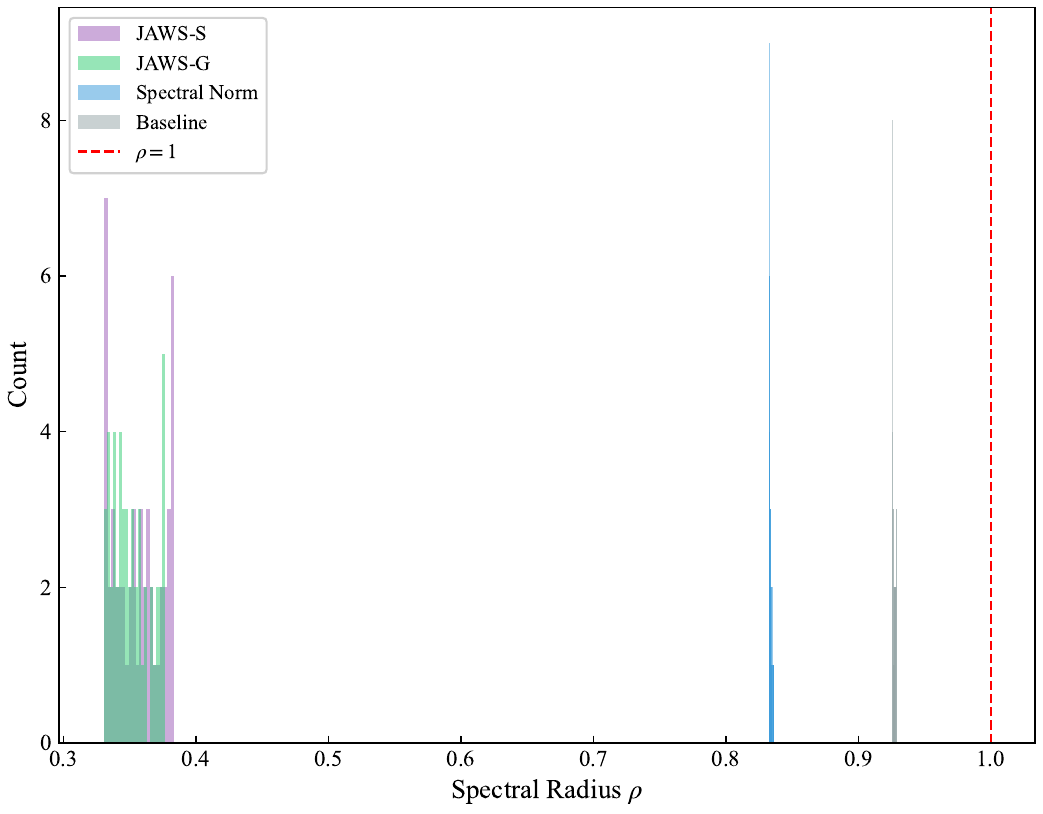}
        \caption{Jacobian Spectral Radius Distribution}
        \label{fig:1d_lyapunov}
    \end{subfigure}
    \caption{Long-term stability and Jacobian eigenvalue analysis (1D Burgers). JAWS-S controls error propagation within a safe range, effectively preventing spectral blow-up (as in the unconstrained Baseline) or excessive dissipation (as in Spectral Normalization).}
    \label{fig:1d_longterm}
\end{figure}

The core challenge for autoregressive models during time-stepping lies in preventing minute prediction errors from being progressively amplified step by step ($\boldsymbol{\epsilon}_{t+1} \approx \mathbf{J} \boldsymbol{\epsilon}_t$), without sacrificing physical features such as shock fronts. Figure~\ref{fig:1d_longterm} illustrates the error distribution after 200 continuous rollout steps.

For models lacking strict temporal dynamic constraints, they tend to fall into two extremes: either, like the unconstrained Baseline, errors progressively accumulate and eventually trigger spectral blow-up; or, like Spectral Normalization, they are forced to learn an overly smooth mapping to avoid divergence, at the cost of severe numerical dissipation. Through analysis of the Jacobian spectrum and wavenumber energy spectrum, we reveal the underlying mechanism.

\begin{figure}[htbp]
    \centering
    \includegraphics[width=0.55\linewidth]{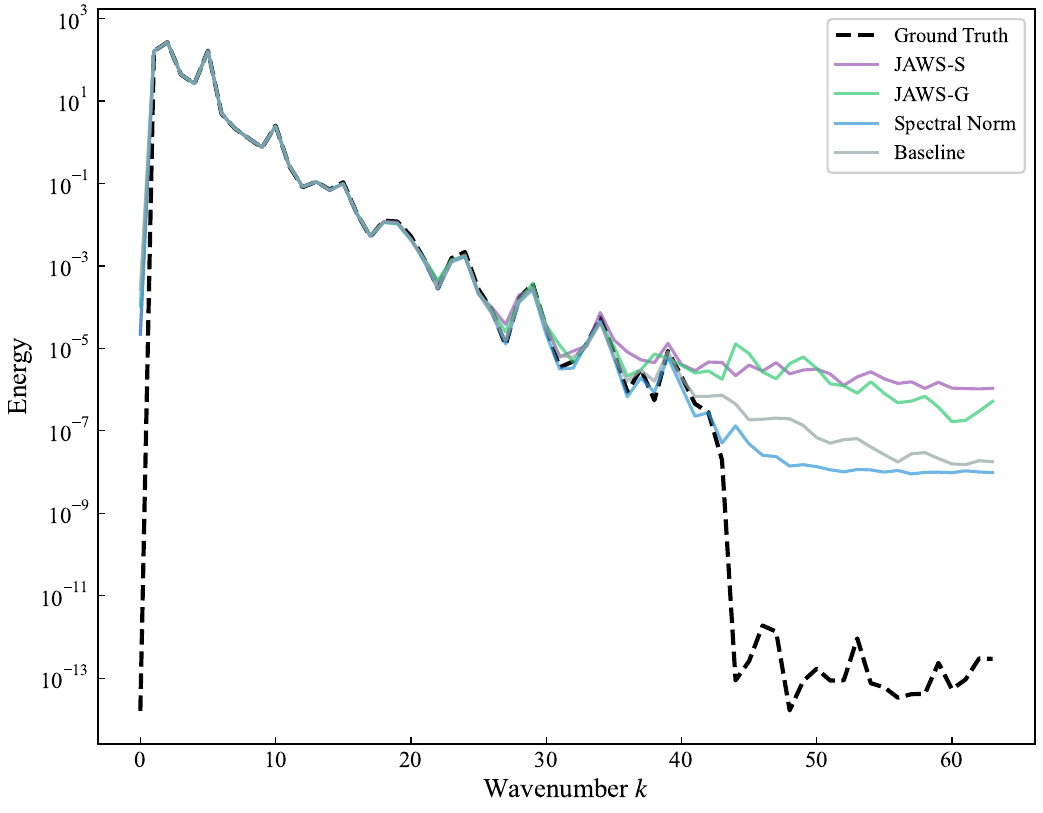}
    \caption{Wavenumber energy spectrum. Baseline and Spectral Normalization barely maintain stability by sacrificing high-frequency features (over-smoothing); JAWS-S, through its spatially-adaptive mechanism, proactively releases high-frequency energy to delineate sharp shocks while relying on the overall Jacobian constraint to firmly lock it at a safe magnitude, successfully breaking the ``smoothness vs.\ accuracy'' seesaw.}
    \label{fig:1d_wavenumber}
\end{figure}

Figure~\ref{fig:1d_wavenumber} reveals a highly counter-intuitive yet physically consistent phenomenon: in the high-frequency range (wavenumber $k>40$), the predicted energy of all neural network models exhibits a flat truncation ``tail'' above the ground truth. It should be noted that, due to the logarithmic scale of the plots, this high-frequency energy that appears to deviate from the ground truth ($10^{-6}$ to $10^{-8}$ magnitude) is actually negligible relative to the total system energy ($10^2$ magnitude)---it merely represents the noise floor arising from the neural network's limited expressiveness and floating-point precision. However, the subtle height differences of this noise floor directly reflect the ``stringency'' of constraints applied to each model: the noise floor of Spectral Normalization and Baseline is strongly suppressed to the lowest level ($\sim 10^{-8}$), which is precisely the result of their indiscriminate smoothing constraints---to forcibly suppress error accumulation, they indiscriminately flatten all fluctuations in the high-frequency range (including the essential high-frequency features constituting the shock front), causing severe dissipation of the flow field's physical structure.

In contrast, JAWS-S maintains the relatively highest noise floor energy level in the high-frequency tail ($\sim 10^{-6}$). This is not predictive divergence, but rather a manifestation of JAWS-S's spatially-adaptive mechanism at work: upon detecting the shock location, it proactively relaxes the Jacobian penalty in the local region, allowing the model to freely retain sufficient high-frequency energy to sharply delineate the shock front. This slight elevation of the noise floor is precisely the cost of avoiding ``over-smoothing.'' Thanks to the overall spectral radius constraint we designed (compressed to the safe interval of $\rho \approx 0.35$), JAWS-S can strictly control this high-frequency energy retained for shock capturing at an absolutely safe, minuscule level, thereby achieving the balance of ``both clearly capturing transient abrupt changes and ensuring stable 200-step long-term rollout.''

\subsubsection{Ablation Analysis: Adaptive Spatial Shock-Capturing Mechanism}
\label{subsubsec:1d_spatial}
\begin{figure}[htbp]
    \centering
    \begin{subfigure}[c]{0.48\linewidth}
        \centering
        \includegraphics[width=\linewidth]{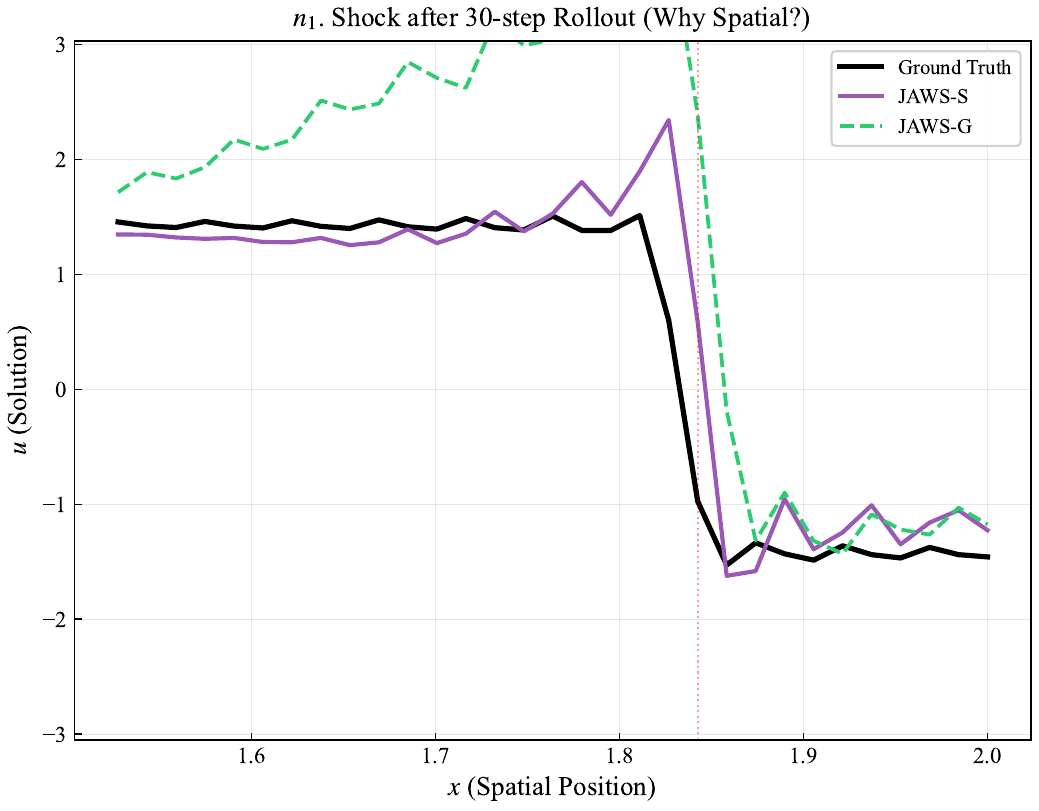}
        \caption{Cross-section at the Singularity}
        \label{fig:1d_shock_cross}
    \end{subfigure}
    \hfill
    \begin{subfigure}[c]{0.48\linewidth}
        \centering
        \includegraphics[width=\linewidth]{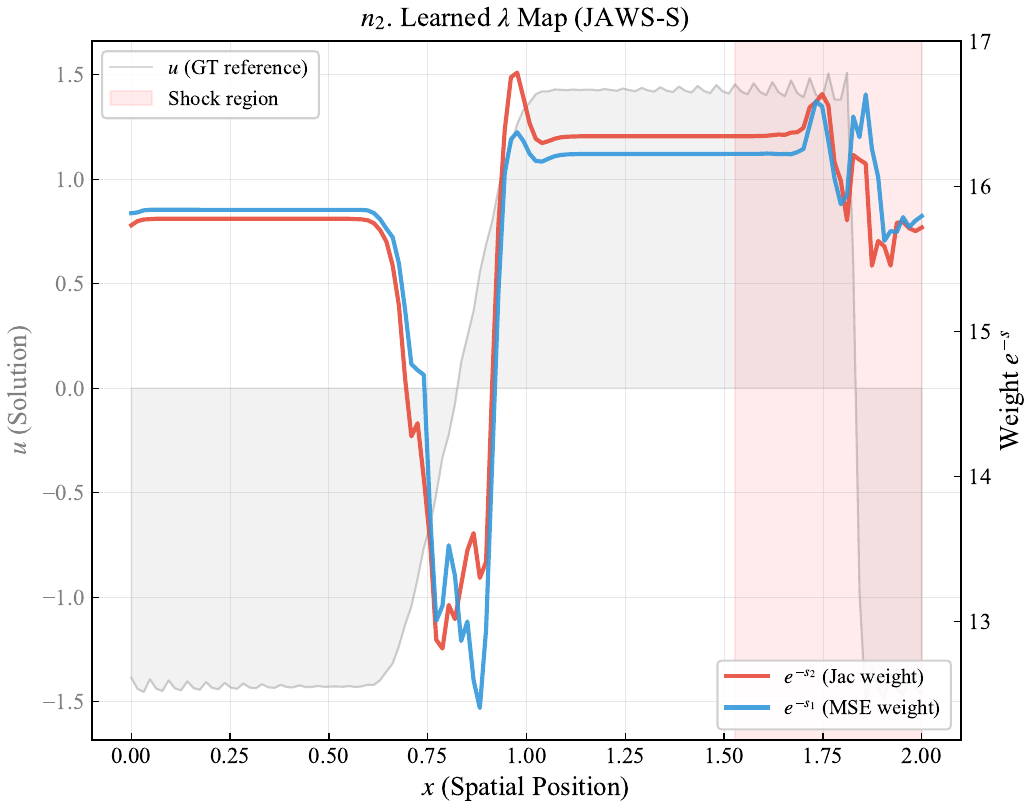}
        \caption{Learned Spatial Weights}
        \label{fig:1d_spatial_map}
    \end{subfigure}
    \caption{Spatially-adaptive mechanism. The model naturally learns to precisely reduce regularization strength (a drop in $e^{-s_2}$) near shock regions to prevent over-smoothing.}
    \label{fig:1d_weight_map}
\end{figure}

To explain how JAWS addresses the problem of ``stability requires smoothing, while clarity risks divergence,'' we perform an ablation analysis by comparing the global variant (JAWS-G) with the spatially-adaptive variant (JAWS-S). Figure~\ref{fig:1d_weight_map} shows the uncertainty distribution learned by the model across the entire spatial domain. As theoretically expected, the global variant JAWS-G applies uniform strict constraints across the entire flow field, causing all high-frequency details (including shock fronts) to be indiscriminately smoothed out, confirming that global regularization inevitably leads to blurring of physical boundaries. In contrast, JAWS-S exhibits remarkably strong adaptive capability: at the physical locations where shock fronts actually exist ($x \approx 0.79$), the model automatically and precisely reduces the regularization penalty strength (i.e., decreasing the weight $e^{-s_2}$); simultaneously, in convective regions where the flow changes drastically, it automatically increases its focus on prediction accuracy (i.e., increasing the weight $e^{-s_1}$). The dynamic coordination of these two weights is spiritually aligned with adaptive strategies in traditional CFD---intelligently modulating constraint strength based on local flow features. Regularization here plays a role analogous to ``numerical viscosity'': enforcing strict constraints in smooth regions to suppress error accumulation, while automatically relaxing constraints near shocks to prevent over-smoothing from destroying sharp physical features. Notably, this spatially-adaptive behavior is entirely learned autonomously by the model based on data, without requiring any prior knowledge of shock locations.

\subsubsection{Spatiotemporal Evolution and Synergy with Pushforward Training}
\label{subsubsec:1d_synergy}

\begin{figure}[htbp]
    \centering
    \begin{subfigure}[c]{0.32\linewidth}
        \centering
        \includegraphics[width=\linewidth]{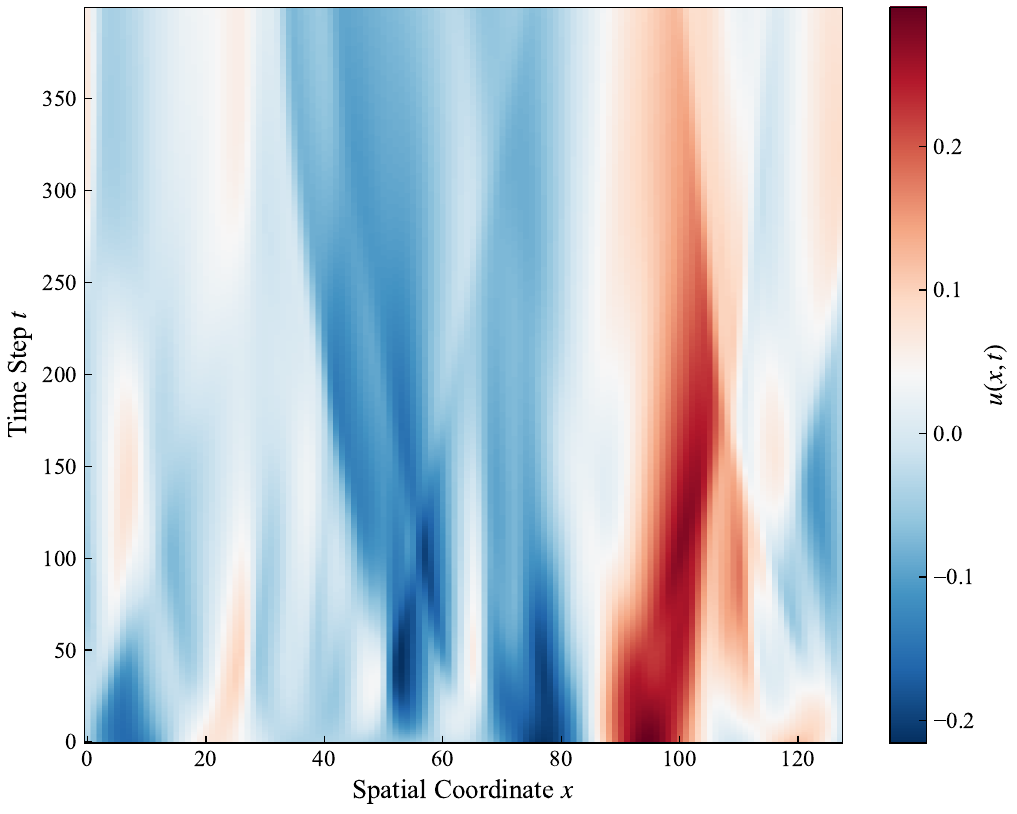}
        \caption{Ground Truth}
        \label{fig:heatmap_gt}
    \end{subfigure}
    \hfill
    \begin{subfigure}[c]{0.32\linewidth}
        \centering
        \includegraphics[width=\linewidth]{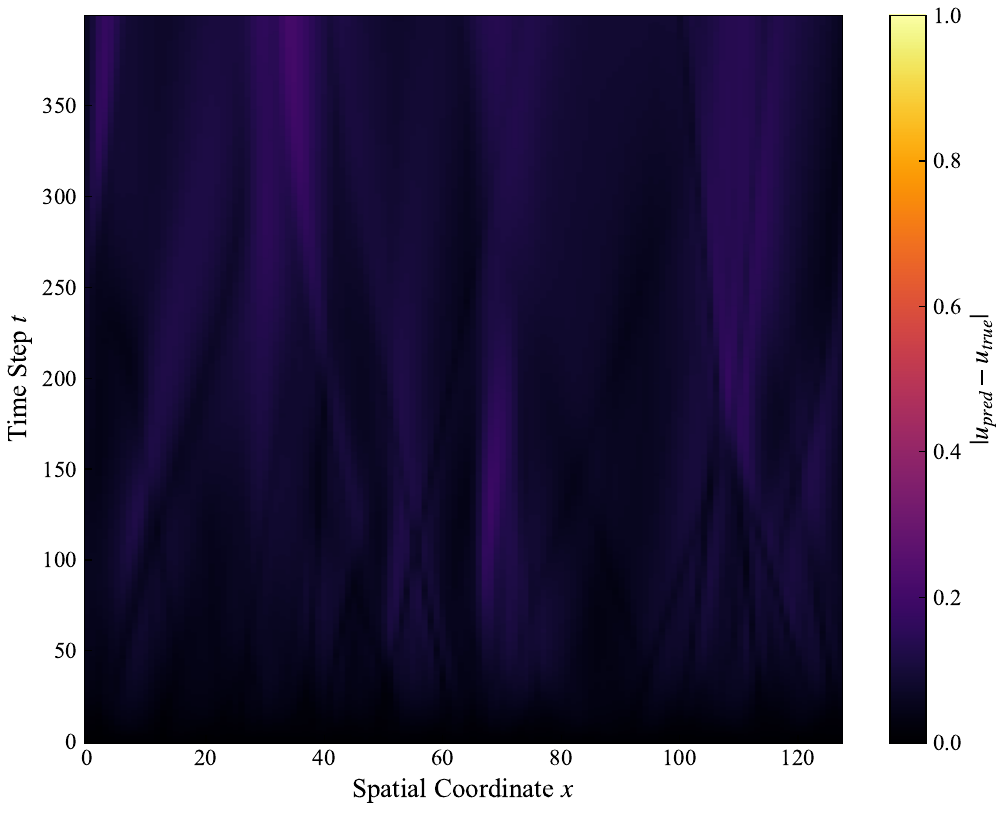}
        \caption{PF-10}
        \label{fig:heatmap_pf10}
    \end{subfigure}
    \hfill
    \begin{subfigure}[c]{0.32\linewidth}
        \centering
        \includegraphics[width=\linewidth]{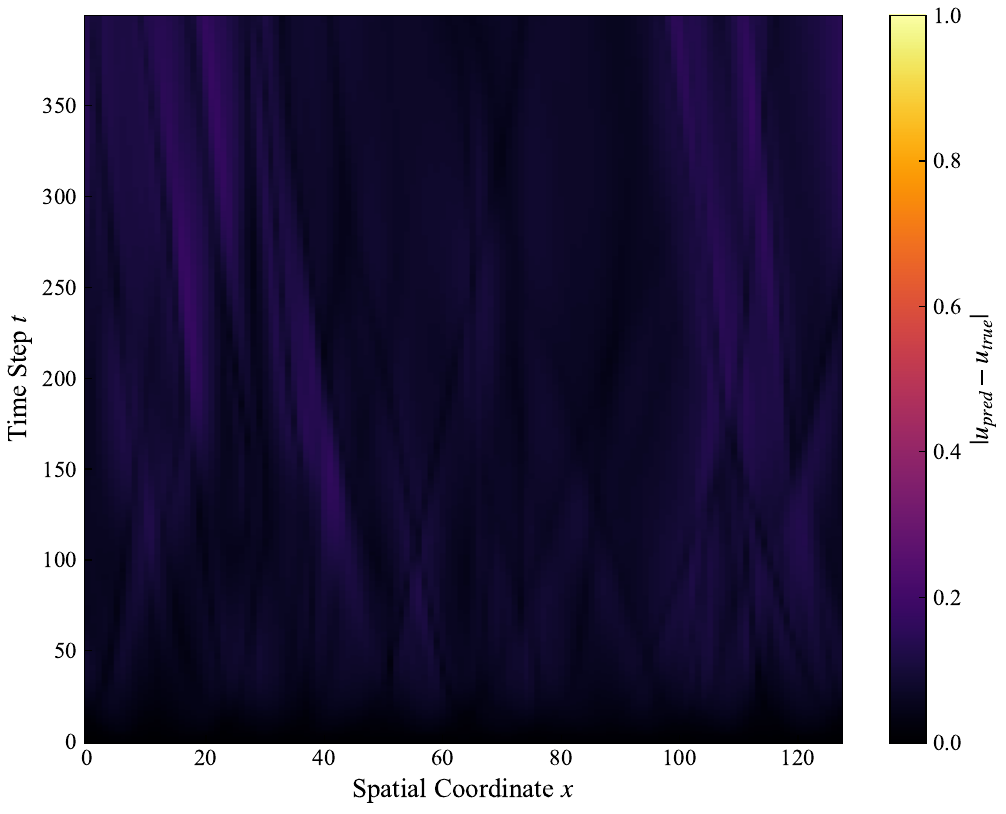}
        \caption{JAWS+PF(5)}
        \label{fig:heatmap_jaws_pf5}
    \end{subfigure}
    \caption{Spatiotemporal error analysis during testing. (a) The ground truth clearly shows the moving shock front. (b) The PF-10 method generates severe local error hotspots along the shock trajectory. (c) The hybrid model JAWS+PF(5) significantly suppresses these error spikes, rendering the error distribution much more uniform across the field.}
    \label{fig:1d_spatiotemporal}
\end{figure}

Combining JAWS with Pushforward training (PF, i.e., time-trajectory window optimization) significantly enhances the model's training efficiency. On the 1D benchmark, the hybrid combination of \textbf{JAWS+PF(5)} achieves an optimal balance among training time, peak memory consumption, and final long-sequence prediction accuracy. The visualization of the spatiotemporal evolution in Figure~\ref{fig:1d_spatiotemporal} intuitively confirms this complementary effect. Models relying purely on long-horizon Pushforward training (PF-10) attempt to brute-force memorize the entire long trajectory, which instead causes errors to aggressively aggregate and cascade near the shock front. Acting as a ``pre-conditioner,'' JAWS effectively pacifies these localized high-frequency anomalous perturbations. Consequently, the model only needs to be unrolled for very short time steps ($k=5$) during training to keep the overall error uniformly bounded across the domain, without imposing additional memory burdens.

(Note: For detailed analyses on initialization stability, rigorous Pareto benchmarking of regularization parameters, and model robustness under explicit Gaussian noise inputs, please refer to Appendix~\ref{sec:appendix_1d}.)

\subsection{2D Flow Past a Cylinder (Re=400)}
\label{subsec:2d_ns}

To further validate the universality and reliability of JAWS in high-dimensional complex flow fields---especially its performance in the face of intense nonlinear transients and complex vortex shedding phenomena---we evaluate its out-of-distribution (OOD) generalization capabilities on the 2D flow past a cylinder scenario.

\subsubsection{Setup and Computational Efficiency}
\textbf{Configuration:} The model is first trained conventionally on a mixed dataset comprising relatively lower Reynolds numbers ($\text{Re} \in \{100, 200, 300\}$). During testing, the conditions are strictly extrapolated to $\text{Re}=400$. At this Reynolds number, the flow develops much more vigorous von K\'{a}rm\'{a}n vortex streets that are unseen in the training set. In testing, the model must perform 200 continuous steps of autoregressive prediction (sampled at 5 Hz, equivalent to a physical time span of 40 seconds) to rigorously examine the accumulation of errors over time. In this experiment, all compared methods uniformly adopt Feature-wise Linear Modulation UNet (FiLM-UNet) as the backbone architecture.

\textbf{FiLM Layer OOD Conditional Input Details:} To achieve continuous generalization across different fluid states, the kinematic viscosity $\nu$ (inversely proportional to $Re$) is input to the network as a global scalar condition. Specifically, this scalar is processed through a Multi-Layer Perceptron (MLP) mapping network to generate the affine scaling and shift parameters for FiLM layers distributed throughout the UNet backbone. During out-of-distribution testing at $Re = 400$, the network infers these affine transformations based on the continuous mapping learned from the training domain ($Re \in \{100, 200, 300\}$). For a detailed discussion of how this FiLM-based dynamic physical scaling mechanism replaces traditional static normalization to preserve physical scale differences, see Appendix~\ref{subsec:app_film}.

\textbf{Computational Cost Analysis:} Table~\ref{tab:re400_stats} lists the computational overhead and final accuracy of different models. Brute-force extension of the traditional Pushforward training window to $\text{PF}(k=10)$ not only causes memory usage to skyrocket (23.44 GB) but also results in exceptionally long training times (over 110 minutes). Spectral Normalization with a $k=5$ window, while close to PF($k=5$) in memory and training time, still has an error of $0.431$ at step 200 with limited improvement. In contrast, JAWS+PF(5) maintains reasonable memory (14.22 GB) and training time, while reducing the relative $L^{2}$ error at step 200 ($0.106$) by approximately $3.5\times$ compared to the next best method (PF($k=10$) at $0.374$).

\begin{table}[htbp]
    \centering
    \caption{Computational efficiency and long-term prediction accuracy for 2D flow past a cylinder OOD rollout (step 200 at $\text{Re}=400$). RMSE denotes Root Mean Square Error.}
    \label{tab:re400_stats}
    \small
    \begin{tabular}{lcccc}
        \toprule
        \textbf{Model} & \textbf{Time (min)} & \textbf{Memory (GB)} & \textbf{Rel-$L^2$ (Step 200)} & \textbf{RMSE} \\
        \midrule
        Baseline (FiLM-UNet) & \textbf{7.59} & \textbf{3.02} & 0.767 & 0.758 \\
        PF($k=5$) & 61.78 & 12.26 & 0.707 & 0.699 \\
        PF($k=10$) & 113.15 & 23.44 & 0.374 & 0.369 \\
        Spectral Norm + PF(5) & 62.30 & 10.15 & 0.431 & 0.426 \\
        JAWS-S + PF(5) & 76.41 & 14.22 & \textbf{0.106} & \textbf{0.104} \\
        \bottomrule
    \end{tabular}
\end{table}

\subsubsection{Error Control and Wake Structure Retention}
\begin{figure*}[htbp]
    \centering
    \includegraphics[width=0.95\linewidth]{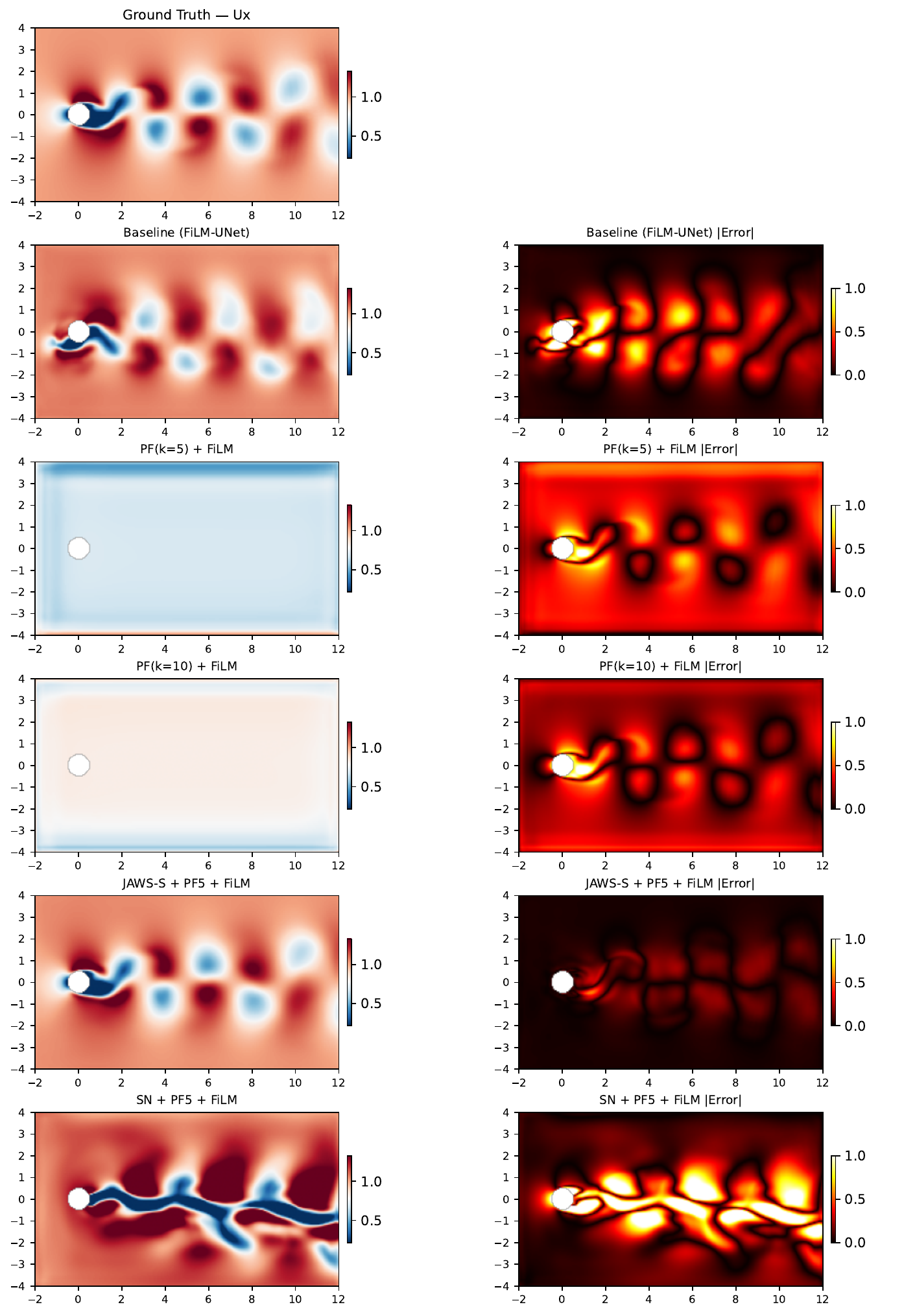}
    \caption{Velocity magnitude (top) and absolute error (bottom) after 200 continuous rollout steps under OOD conditions ($\text{Re}=400$). JAWS uniquely prevents the degradation of flow structures.}
    \label{fig:re400_snapshot}
\end{figure*}

To visually demonstrate the consequences of lacking temporal constraints, we extract the velocity magnitude fields ($|U|$) at step 200 and their corresponding absolute error maps (Figure~\ref{fig:re400_snapshot}). For models relying solely on local MSE loss for training, errors continuously aggregate and amplify in the wake region, ultimately blurring the physical contours beyond recognition. Thanks to JAWS's stable modulation of the spectrum throughout the entire prediction process, the model is able to maintain the structural features of the downstream wake with remarkable clarity, confining errors to an extremely narrow range.

\subsubsection{Energy Conservation and Macroscopic Physical Statistics}
Relying merely on point-wise errors is insufficient to comprehensively measure a model's ability to simulate nonlinear fluid systems. Traditional unconstrained models often struggle to resist numerical dissipation, leading to artificial decay of system energy.

\begin{figure}[p]
    \centering
    \includegraphics[width=0.85\linewidth]{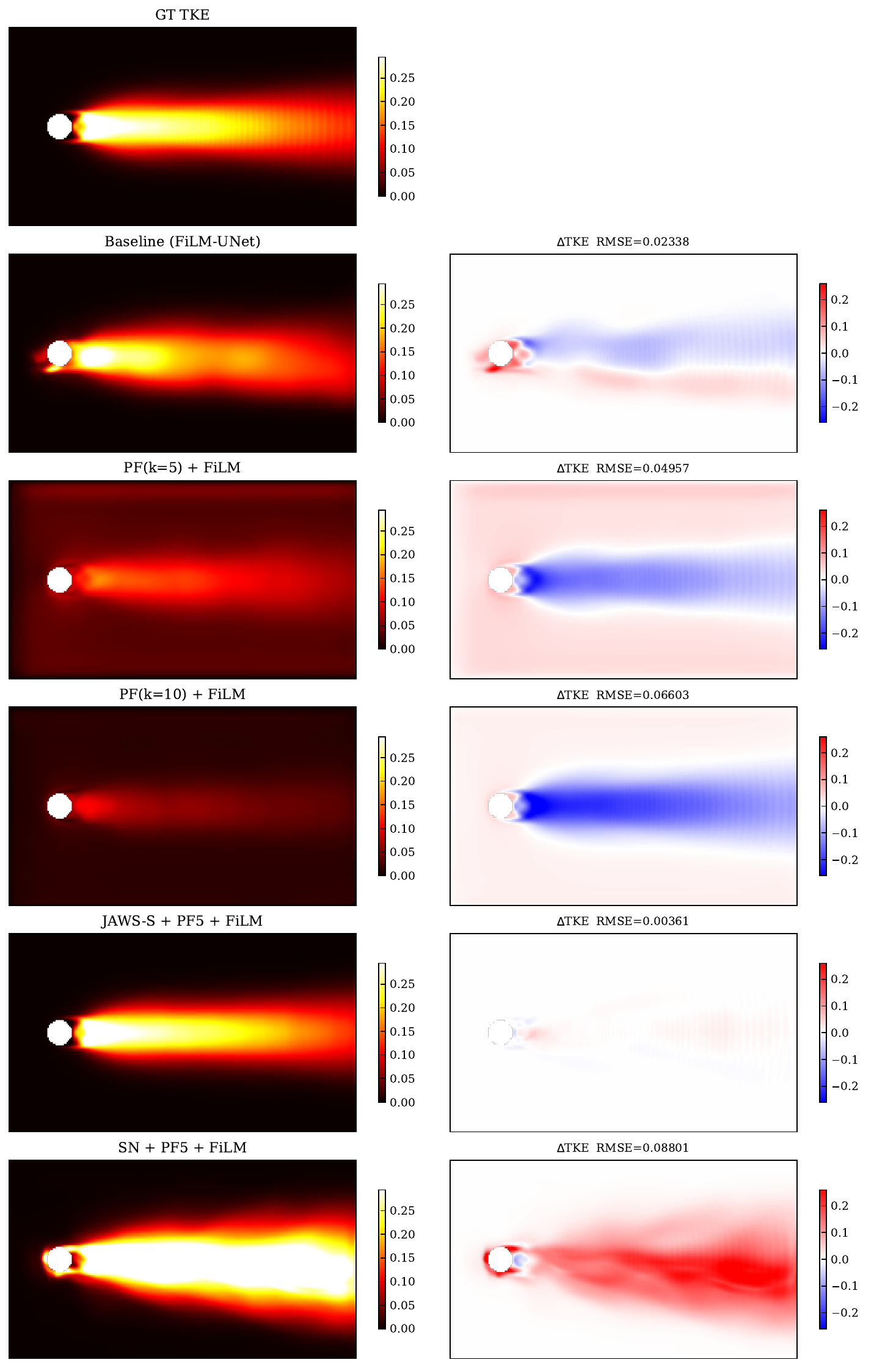}
    \caption{Turbulent kinetic energy distribution, demonstrating the model's ability to preserve internal energy. Models lacking spectral constraints suffer severe energy dissipation.}
    \label{fig:re400_tke}
\end{figure}

\begin{figure}[p]
    \centering
    \includegraphics[width=\linewidth]{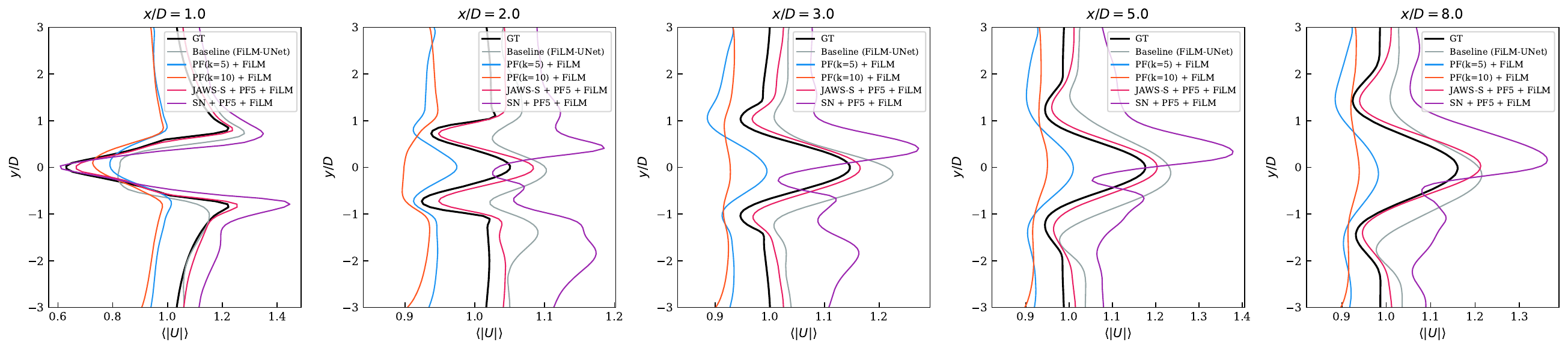}
    \caption{Streamwise velocity profiles at various downstream cross-sections. JAWS perfectly retains momentum distribution in the far wake.}
    \label{fig:re400_wake}
\end{figure}

Therefore, we examine the model's energy retention capabilities by analyzing the Turbulent Kinetic Energy (TKE) and the wake velocity profiles along the streamwise direction ($x/D$). As shown in Figures~\ref{fig:re400_tke} and \ref{fig:re400_wake}, both the Baseline and the traditional multi-step Pushforward framework suffer from severe energy dissipation, completely failing to reproduce the strong TKE features in the downstream vortex shedding region. Similarly, in the far downstream region ($x/D \ge 5.0$), their predicted velocity profiles become abnormally flat, confirming severe momentum decay.

JAWS successfully overcomes systemic numerical dissipation. Its predicted macroscopic energy peaks are statistically highly accurate, closely matching CFD results, and this advantage is maintained even into the far downstream limits ($x/D = 8.0$).

\subsubsection{Attractor Dynamics and Frequency Response}
\label{subsubsec:attractor}

The ultimate standard for validating a fluid dynamics surrogate model is whether it can accurately reproduce the system's intrinsic attractor structure. We plot the temporal evolution of the drag coefficient ($C_d$) against the lift coefficient ($C_l$) as a phase portrait, which intuitively reflects the core limit-cycle characteristics of the system.

\begin{figure*}[p]
    \centering
    \begin{subfigure}[b]{0.32\linewidth}
        \centering
        \includegraphics[width=0.95\linewidth]{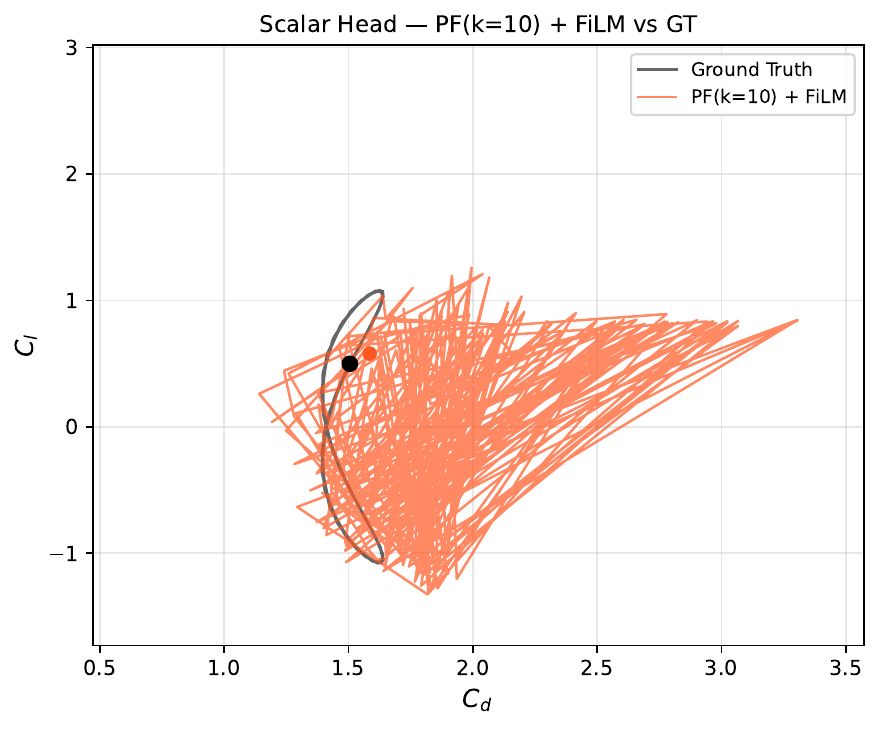}
        \caption{PF($k=10$)}
        \label{fig:phase_pf10}
    \end{subfigure}
    \hfill
    \begin{subfigure}[b]{0.32\linewidth}
        \centering
        \includegraphics[width=0.95\linewidth]{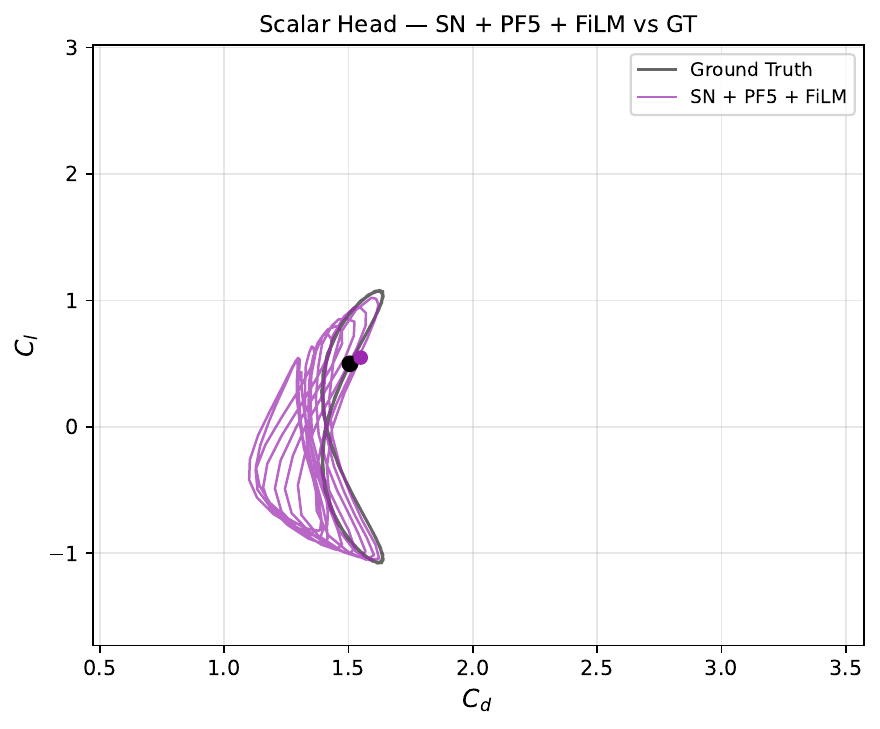}
        \caption{Spectral Norm + PF(5)}
        \label{fig:phase_sn}
    \end{subfigure}
    \hfill
    \begin{subfigure}[b]{0.32\linewidth}
        \centering
        \includegraphics[width=0.95\linewidth]{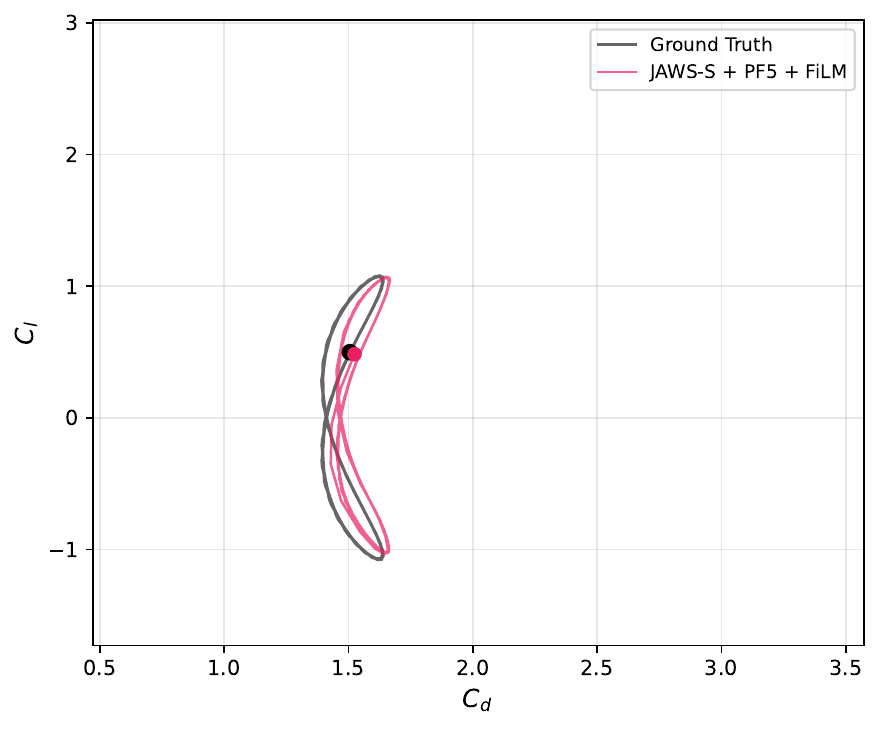}
        \caption{JAWS + PF(5)}
        \label{fig:phase_jaws}
    \end{subfigure}
    \caption{Drag-lift phase portraits ($C_l$ vs. $C_d$) reflecting the limit-cycle dynamics of the system. The unconstrained PF($k=10$) (left) inevitably diverges into chaos, Spectral Normalization (center) shows improvement but still exhibits phase drift, while JAWS (right) accurately locks onto the intrinsic dynamic attractor, maintaining topological stability.}
    \label{fig:re400_phase_portraits}
\end{figure*}

In Figure~\ref{fig:re400_phase_portraits}, due to the lack of spectral stability protection, traditional multi-step Pushforward models (such as PF($k=10$)) progressively destabilize during autoregressive iteration: their phase trajectories fail to maintain the periodicity of the limit cycle, exhibiting not only severe amplitude deviations but ultimately degenerating into a tangle of completely irregular ``chaotic lines.'' JAWS, on the other hand, firmly locks onto the system's topological structure; its predicted phase trajectory closes perfectly and overlaps precisely with the true CFD attractor, introducing no geometric drift.

\begin{figure*}[p]
    \centering
    \includegraphics[width=0.95\linewidth]{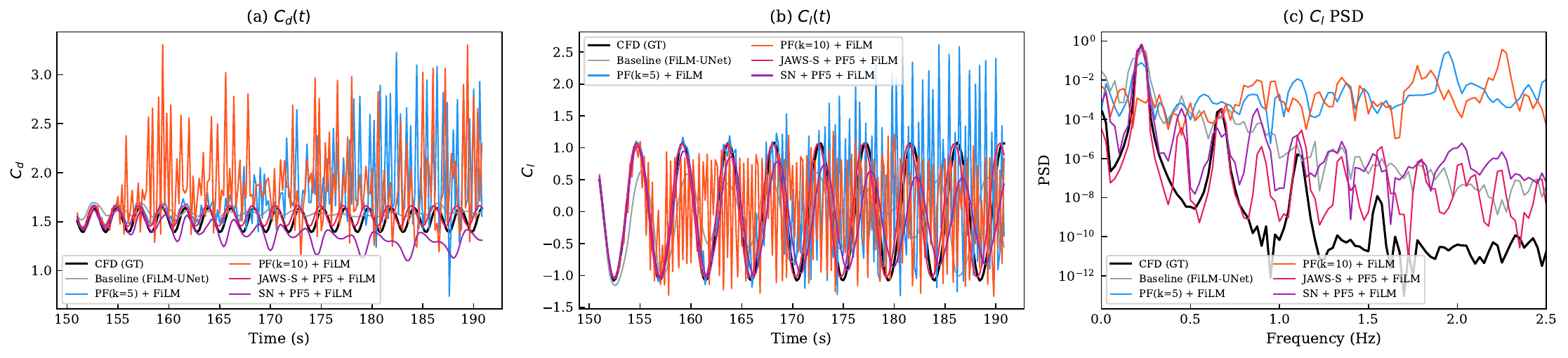}\\
    \vspace{0.3cm}
    \caption{Macroscopic dynamics validation. JAWS accurately captures physical frequencies over long horizons and perfectly reproduces the intrinsic geometric structure of trajectories in 2D flow past a cylinder.}
    \label{fig:re400_attractor}
\end{figure*}

Quantitative analysis of the Power Spectral Density (PSD) (see Figure~\ref{fig:re400_attractor}) further substantiates this: traditional Pushforward models exhibit severe frequency shifting and amplification of spurious frequencies when predicting critical fluid frequencies such as the Strouhal number; in contrast, JAWS displays a precise spectral response that is highly consistent with the true physical periodicities.

\subsubsection{In-depth Analysis of Divergence Failure Modes in Ablation}
\label{subsubsec:divergence_modes}

Comparing the two diametrically opposed failure modes of the unconstrained long-sequence method (PF($k=10$)) and the standalone local spatial regularization (JAWS-S) yields deep physical insights. From the empirical data, although both exhibit comparable stability horizons during the initial unrolling phase, their ultimate degradation paths are diametrically opposed.

The unconstrained PF($k=10$) suffers from progressive spectral divergence. Lacking hard Lyapunov contraction bounds, it allows minute high-frequency aliasing errors to gradually compound through the convective terms, ultimately causing the phase trajectory to spiral away from the physical limit cycle and collapse into chaos (as shown in Figure~\ref{fig:phase_pf10}). Conversely, the standalone JAWS-S model degrades through artificial state-locking. By prioritizing absolute mathematical stability ($\rho(J) \le 1$) over long-range convective fidelity, the standalone model ultimately imposes excessive numerical dissipation globally to compensate for accumulated perturbations. This over-damping effectively ``freezes'' the transient dynamics, leading to premature, unphysical decay of macroscopic kinetic energy.

The hybrid integration (JAWS-S + PF(5)) elegantly resolves this antinomy: by leveraging JAWS to robustly suppress local high-frequency instabilities (acting as the ``brake''), the short-window Pushforward module is completely liberated to focus single-mindedly on steering the low-frequency convective phase (acting as the ``steering wheel''). The results in Table~\ref{tab:re400_stats} quantitatively confirm this synergistic effect---the hybrid method achieves a multi-fold improvement in relative $L^2$ error over the next best method, while keeping computational costs within acceptable bounds.

%% file: sections/05_discussion.tex
\section{Discussion: Attractor Locking Under Quasi-2D Constraints}
\label{sec:discussion}

From the perspective of physical rigor, we must note that the dataset at Reynolds numbers such as $Re = 400$ was generated under strict two-dimensional (2D) computational constraints. Specifically, although the OpenFOAM simulation domain has non-zero thickness in the $z$-direction, the mesh has only a single layer in that direction with \texttt{empty} boundary conditions imposed, which is numerically equivalent to enforcing the solution of the pure 2D Navier-Stokes equations.

From the essential viewpoint of fluid mechanics, beyond $Re \approx 190$, the wake behind the cylinder inevitably undergoes three-dimensional bifurcation (Mode~A and Mode~B instabilities) \cite{williamson1996vortex}. By artificially suppressing the three-dimensional vortex stretching mechanism in the computational domain (i.e., enforcing $\boldsymbol{\omega} \cdot \nabla \mathbf{u} = 0$), our training and extrapolation datasets effectively represent a ``pseudo-physical'' pure 2D Navier-Stokes manifold. In this artificially constrained two-dimensional flow field, the three-dimensional forward energy cascade is blocked, replaced by a forward cascade of enstrophy toward small scales. Due to the absence of genuine three-dimensional dissipation mechanisms, this readily leads to anomalous accumulation of high-frequency vorticity components in the high-wavenumber region.

However, this physical limitation, in an unintended manner, actually highlights the algorithmic robustness of the JAWS method. For standard autoregressive models, this unphysical high-frequency enstrophy accumulation in two-dimensional chaotic flows is precisely the primary trigger for error accumulation and spectral blow-up. Remarkably, despite operating within this purely mathematical 2D system that is prone to numerical divergence, JAWS~+~PF still successfully discovers and enforces the underlying dynamical invariants. By leveraging spatially heteroscedastic uncertainty to locally ``absorb'' this high-frequency numerical and spectral noise, the model stably locks onto the chaotic two-dimensional limit cycle (as shown in Figure~\ref{fig:re400_phase_portraits}). This demonstrates that JAWS is not merely memorizing shallow visual data, but has fundamentally mastered how to stabilize those highly nonlinear PDE mathematical attractors where energy tends to accumulate.

This observation simultaneously provides a clear direction for further extension of the method: when JAWS is extended to genuine three-dimensional turbulence simulations in the future, the natural energy cascade channel provided by three-dimensional vortex stretching should theoretically further \textit{reduce} the risk of trajectory divergence, potentially making JAWS's stability advantages even more pronounced.

%% file: sections/06_conclusion.tex
\clearpage
\section{Conclusion}
\label{sec:conclusion}

This paper proposes JAWS, a probabilistic regularization framework designed to balance stability and fidelity in autoregressive data-driven surrogate models. Unlike uniform Jacobian constraints that trade numerical stability at the cost of over-smoothing physical discontinuities, JAWS utilizes aleatoric uncertainty as a proxy for spatially-adaptive spectral attention. This mechanism empowers the model to learn to relax constraints near steep gradients---facilitating the implementation of shock-capturing behaviors---while enforcing strict contraction in smooth regions to suppress error accumulation.

Experiments demonstrate that JAWS serves as an effective spectral pre-conditioning method for trajectory optimization. By utilizing a gradient detachment strategy, local uncertainty estimation is decoupled from long-term accumulated gradients, enabling memory-efficient, short-horizon optimization windows to focus on correcting low-frequency errors, thereby overcoming the memory limitations inherent in autoregressive training. Evaluations on the 1D Burgers' equation and strongly nonlinear 2D flow past a cylinder verify that JAWS not only guarantees long-term autoregressive stability but also fundamentally preserves critical macroscopic physical features---including turbulent kinetic energy, vortex wake profiles, and the physical accuracy of limit-cycle attractors.

Future work will explore extending this spatially-adaptive regularization method to unstructured grids and 3D turbulence simulation models, further evaluating the role of uncertainty-aware spectral conditioning in the construction of robust scientific machine learning models.

%% file: sections/07_appendix.tex
﻿\section{Appendix for 1D Burgers' Equation Validations}
\label{sec:appendix_1d}

\subsection{Noise Robustness via Aleatoric Uncertainty}
\label{subsec:app_noise}
\begin{figure}[H] 
    \centering
    \includegraphics[width=0.6\linewidth]{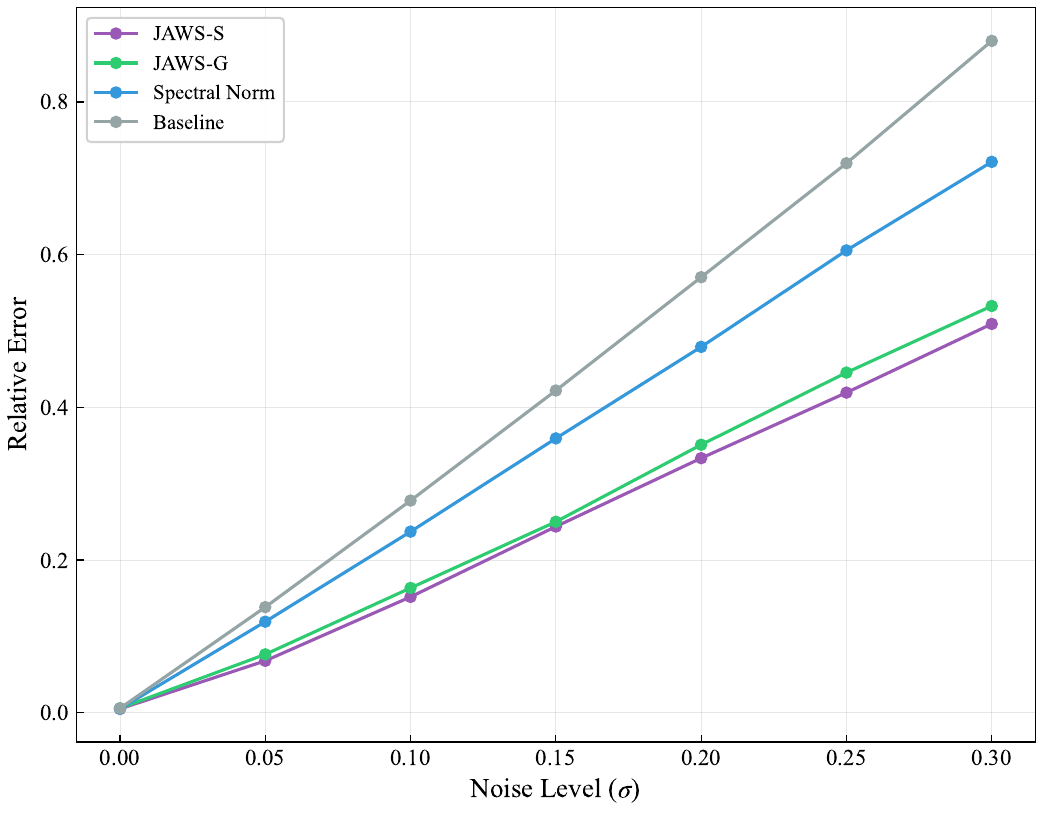}
    \caption{Relative $L^2$ error vs. Input Noise $\sigma$. JAWS variants exhibit a flat error response compared to standard baselines.}
    \label{fig:noise_robustness}
\end{figure}

We evaluate the model's potential for practical deployment by injecting Gaussian noise ($\sigma \in [0, 0.3]$) into the test inputs (Figure~\ref{fig:noise_robustness}). Standard MSE-based models inherently assume infinite data precision, making them highly susceptible to overfitting high-frequency Gaussian noise. In contrast, JAWS exhibits exceptional robustness due to the Bayesian NLL loss, where the learned precision acts as an adaptive Signal-to-Noise Ratio (SNR) estimator. 

\subsection{Out-of-Distribution (OOD) Initialization Generalization}
\label{subsec:app_ood}
Table~\ref{tab:ood} reports OOD behaviors (low/high viscosity and variable frequencies) based strictly on single-step extrapolations. Unconstrained physics simulators intuitively accomplish lowest standalone MSE deviations, but sacrifice absolute recursive integrity demonstrated in the primary text.

\begin{table}[H]
    \centering
    \caption{Relative $L^2$ error on 1D OOD test scenarios (single-step calculations). Models prioritizing temporal invariances exhibit slight localized performance tradeoffs but dominate holistic evaluation timelines.}
    \label{tab:ood}
    \small
    \begin{tabular}{lcccc}
        \toprule
        \textbf{Model} & \textbf{Low $\nu$} & \textbf{High $\nu$} & \textbf{High Freq} & \textbf{Interp.} \\
        \midrule
        JAWS-S      & 0.036 & 0.008 & 0.162 & 0.004 \\
        JAWS-G      & 0.031 & 0.008 & 0.155 & 0.004 \\
        Spec.\ Norm & 0.027 & 0.008 & \textbf{0.048} & 0.004 \\
        Baseline    & \textbf{0.009} & 0.008 & 0.053 & \textbf{0.003} \\
        \bottomrule
    \end{tabular}
\end{table}

\subsection{Convexity and Initialization Robustness}
\label{subsec:app_convexity}
\begin{figure}[H]
    \centering
    \includegraphics[width=0.6\linewidth]{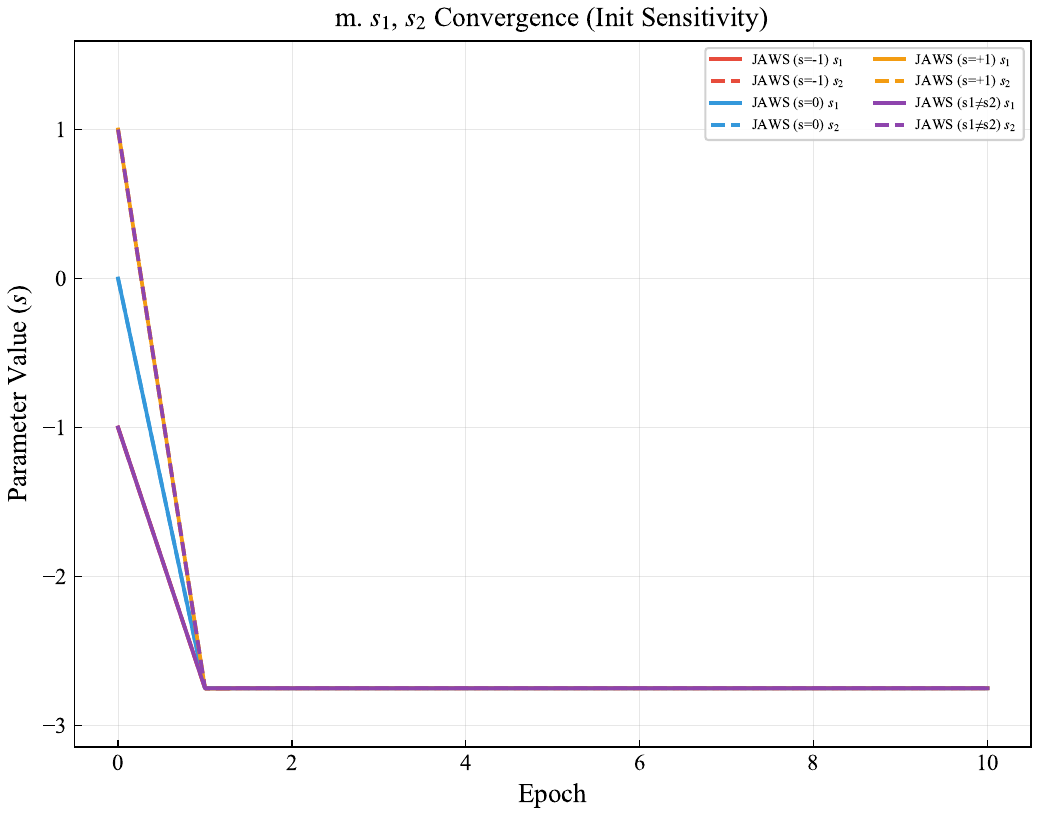}
    \caption{Convergence of uncertainty parameters $s_1, s_2$ verifying underlying loss landscape convexity.}
    \label{fig:ablation_c}
\end{figure}

To assess mathematical rigor behind autonomous parameter searching, Figure~\ref{fig:ablation_c} visualizes varied initialization origins for scalar uncertainties $(s_1, s_2)$ across training steps. They deterministically converge towards identical optimal values, proving strictly convex topological loss landscapes negating manual tuning constraints entirely.

\section{Practical Optimization Strategies and Architectural Inductive Biases}
\label{sec:appendix_engineering}

To successfully scale the theoretical Maximum A Posteriori estimation to high-dimensional, strongly nonlinear fluid dynamics applications, we implement several critical architectural inductive biases and optimization strategies. These are not mere engineering compromises, but rather core topological designs that bridge continuous mathematical theory and discrete neural network optimization.

\subsection{Structured Information Bottleneck: Asymmetric Uncertainty Parameterization}
\label{subsec:app_bottleneck}

Although Equation~(\ref{eq:jaws_final}) idealizes both $s_1$ and $s_2$ as spatially heteroscedastic fields, granting both full spatial degrees of freedom during training would render the loss landscape highly non-convex, readily causing the model to collapse into trivial solutions---for instance, the model would globally and artificially inflate $s_1(\mathbf{x})$ to escape fitting complex dynamical features. To enforce a structured information bottleneck, we deliberately restrict $s_1$ to a single globally learnable scalar ($s_1 \in \mathbb{R}$), while retaining only $s_2(\mathbf{x})$ as the spatially-adaptive field. This asymmetry forces the base PDE solver to strictly fit the global temporal evolution, while reserving the privilege of ``spatial relaxation'' exclusively for the Jacobian penalty term.

\subsection{Feature-Level Gradient Topology Isolation}
\label{subsec:app_gradient_isolation}

In our dual-head architecture, the local uncertainty field $s_2(\mathbf{x})$ is generated by an auxiliary network $\mathcal{H}_\phi$. Crucially, the hidden features extracted by the base UNet must be explicitly detached before being input to $\mathcal{H}_\phi$. Without this feature-level explicit gradient isolation, the gradients generated by minimizing the regularization penalty term ($0.5\,s_2$) would backpropagate into the backbone UNet, producing ``parasitic gradients'' irrelevant to prediction accuracy. This topological isolation mechanism ensures that the base operator's representation learning is driven entirely by genuine physical reconstruction errors, preventing the auxiliary Jacobian constraints from contaminating the physical manifold of the fluid.

\subsection{FiLM-Based Dynamic Physical Scaling}
\label{subsec:app_film}

Traditional data-driven surrogate models typically rely on static global Z-score normalization, which inadvertently destroys the inherent physical scale differences across different fluid states. To achieve continuous out-of-distribution (OOD) generalization to Reynolds number $Re=400$, we discard global normalization and traditional BatchNorm affine parameters. Instead, we feed the kinematic viscosity (inversely proportional to $Re$) into a Multi-Layer Perceptron (MLP) to generate the affine transformation parameters ($\gamma, \beta$) for Feature-wise Linear Modulation (FiLM) layers. This enables the surrogate model to learn a dynamic physical scaling mechanism that adaptively modulates the feature distribution of the manifold based on macroscopic control parameters.